\documentclass[lettersize,journal]{IEEEtran}
\usepackage{amsmath,amsfonts}
\usepackage{array}
\usepackage[caption=false,font=normalsize,labelfont=sf,textfont=sf]{subfig}
\usepackage{textcomp}
\usepackage{stfloats}
\usepackage{url}
\usepackage{verbatim}
\usepackage{graphicx}
\def\BibTeX{{\rm B\kern-.05em{\sc i\kern-.025em b}\kern-.08em
    T\kern-.1667em\lower.7ex\hbox{E}\kern-.125emX}}
\usepackage{balance}
\usepackage{multirow}
\usepackage{makecell}
\usepackage{xcolor}
\usepackage{colortbl,booktabs}
\usepackage{algorithmicx,algorithm,algpseudocode}
\usepackage{threeparttable}
\usepackage{amssymb}
\usepackage{utfsym}

\begin{document}
\title{MFDS-Net: Multi-Scale Feature Depth-Supervised Network for Remote Sensing Change Detection with Global Semantic and Detail Information}
\author{Zhenyang Huang, Zhaojin Fu, Song Jintao, Genji Yuan, Jinjiang Li  
\thanks{H. Zhen and is with School of Information and electronic engineering, Shandong Technology and Business University, Y antai 264005, China (e-mail: huazhen@sdtbu.edu.cn)} 
\thanks{J. Song, G. Yuan and J. Li are with School of Computer Science and Technology, Shandong Technology and Business University, Y antai 264005, China}}

\markboth{Journal of \LaTeX\ Class Files,~Vol.~18, No.~9, May~2023}%
{How to Use the IEEEtran \LaTeX \ Templates}

\maketitle

\begin{abstract}
Change detection as an interdisciplinary discipline in the field of computer vision and remote sensing at present has been receiving extensive attention and research. Due to the rapid development of society, the geographic information captured by remote sensing satellites is changing faster and more complex, which undoubtedly poses a higher challenge and highlights the value of change detection tasks. We propose MFDS-Net: Multi-Scale Feature Depth-Supervised Network for Remote Sensing Change Detection with Global Semantic and Detail Information (MFDS-Net) with the aim of achieving a more refined description of changing buildings as well as geographic information, enhancing the localisation of changing targets and the acquisition of weak features. To achieve the research objectives, we use a modified $\text{Re}sNet_{34}$ as backbone network to perform feature extraction and DO-Conv as an alternative to traditional convolution to better focus on the association between feature information and to obtain better training results. We propose the Global Semantic Enhancement Module (GSEM) to enhance the processing of high-level semantic information from a global perspective. The Differential Feature Integration Module (DFIM) is proposed to strengthen the fusion of different depth feature information, achieving learning and extraction of differential features. The entire network is trained and optimized using a deep supervision mechanism. 

The experimental outcomes of MFDS-Net surpass those of current mainstream change detection networks. On the LEVIR dataset, it achieved an F1 score of 91.589 and IoU of 84.483, on the WHU dataset, the scores were F1: 92.384 and IoU: 86.807, and on the GZ-CD dataset, the scores were F1: 86.377 and IoU: 76.021. The code is available at https://github.com/AOZAKIiii/MFDS-Net

\end{abstract}


\section{Introduction}
\IEEEPARstart{I}n the field of remote sensing, change detection focuses on information captured by satellites about changes in an area over a period of time. With the upgrading of image acquisition equipment and more specialised domain classification, satellites \cite{1,2,3} (e.g. Landsat9, Gaofen3, etc.) targeting different geographical areas and functions have been applied, expanding the dataset while expanding the range of applications for change detection, such as land cover change \cite{4}, forest change \cite{5}, disaster assessment \cite{6}, urban development \cite{7,8}, etc. Traditional algorithms have given birth to many better algorithms \cite{9,10,11} in the field of change detection, which have made important contributions to the change detection task as well as to the analysis of remote sensing data. Although traditional algorithms often contain noise in their processing results and often do not show extensive advantages in the face of complex samples, their processing ideas still provide an important reference value for deep learning. The emergence of deep learning makes it easier for CD tasks to cope with complex environments and complex change information, thus effectively identifying change targets. Convolutional neural networks (CNNs) have achieved many excellent results \cite{12,13,14} in CD tasks and other fields \cite{xu2022morphtext,xu2022semantic}.

However, we find through extensive experiments that current mainstream methods are still deficient in areas such as edge feature processing of changing targets and sample processing of colour diversity. This problem is mainly caused by a combination of the complexity of the CD data and the limitations of the method itself.

In terms of data, there are still significant differences between samples in the CD dataset. For samples obtained in residential areas, there are similar in shape and size of the changed targets, while for samples obtained in commercial and industrial areas, there are changed targets with diverse colors and complex structures. Due to the different acquisition times and seasons of bi-temporal images, there are significant differences between the environments where changed targets are located and the environments before the changes. In addition, some samples contain a large number of changed targets, while some samples contain only a few or even no changed targets. These factors to some extent increase the difficulty of model processing.

In terms of methods, many methods lack attention to detail information, which can better recover target contours and edges during the reconstruction phase of change detection. Some methods enhance the overall performance of the network by strengthening its ability to focus on global features. However, during the calculation of global feature correlations, environmental noise factors are also taken into account, which causes the network to perform poorly when dealing with complex samples.

For example, IFNet \cite{15} creates a deep supervision network that uses VGG \cite{16} as the backbone network for feature extraction. Subsequently, the difference discrimination network (DDN) performs difference feature extraction and sets accompanying outputs to participate in loss calculation for processing layers of different scales, which enhances the training process. Furthermore, the DDN uses a concatenation of channel attention module and spatial attention module \cite{17} to amplify the focus on feature information. Nonetheless, VGG fails to capture the interrelations among features on a global scale, IFNet has low recognition ability for environmental noise, which leads to unsatisfactory performance in handling fine-grained features. ChangeFormer \cite{18} uses Transformer \cite{19} to create a backbone network and complete feature extraction, capturing long-distance dependency relationships between feature information through self-attention modules. ChangeFormer also creates a Difference Module to process the skip connections of each layer and continuously passes the obtained processing results to the lower layers of the network, thereby continuously strengthening the intensity of the difference features. Although the backbone network uses self-attention to associate global feature information, this also leads to a higher computational cost, and more importantly, the environment features unrelated to the target are also modeled, which often prevents the network from showing processing advantages when facing complex samples.

\begin{figure}
	\begin{center}
		\begin{tabular}{c}
			\includegraphics[height=4cm]{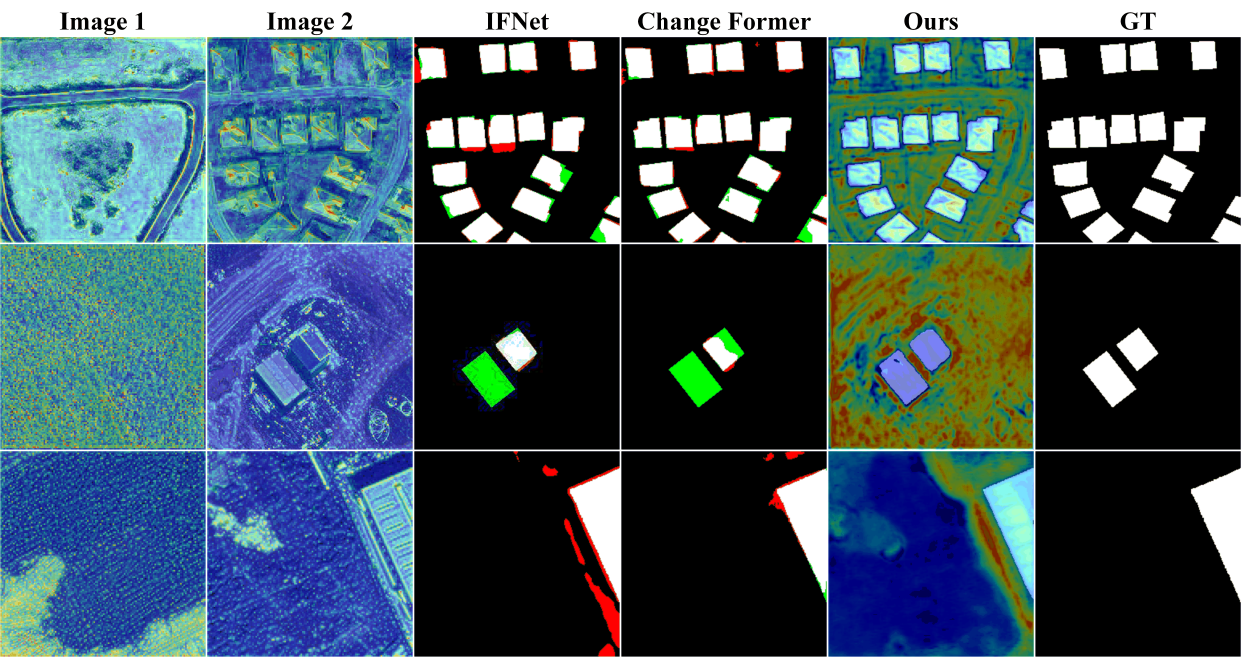}
		\end{tabular}
	\end{center}
	\caption
	{ \label{FIG:1}
		The results of the experiments performed on three datasets, the first sample set is the LEVIR-CD dataset, the second sample set is the WHU-CD dataset and the third is the GZ-CD dataset. The last column shows the heat map obtained from Ours. }
\end{figure}

To address the challenges posed by the complexity of the current CD dataset and to overcome limitations in current methods, we propose MFDS-Net. Taking into account the issue of computation complexity, DO-Conv is used to replace traditional convolutions in MFDS-Net. To address the issue of insufficient attention to detail features, MFDS-Net proposes a Multi-scale Detail Preservation Module (MDPM). It enhances the intensity of local detail information by combining high-frequency information, densely connected convolutional blocks, and residual connections. Through expanding the field of view, it establishes broader correlations among local feature information. To address the issue of global semantic information, MFDS-Net proposes GSEM, which enhances the channels where the target resides while suppressing irrelevant environmental information and models the long-range dependency between feature information using Non-local blocks, thus improving MFDS-Net's capacity to attend to feature information. To better focus on changing targets during the target reconstruction phase, MFDS-Net also proposes DFIM. In the training process, MFDS-Net utilizes a deep supervision mechanism.

\begin{figure*}
	\begin{center}
		\begin{tabular}{c}
			\includegraphics[height=4cm]{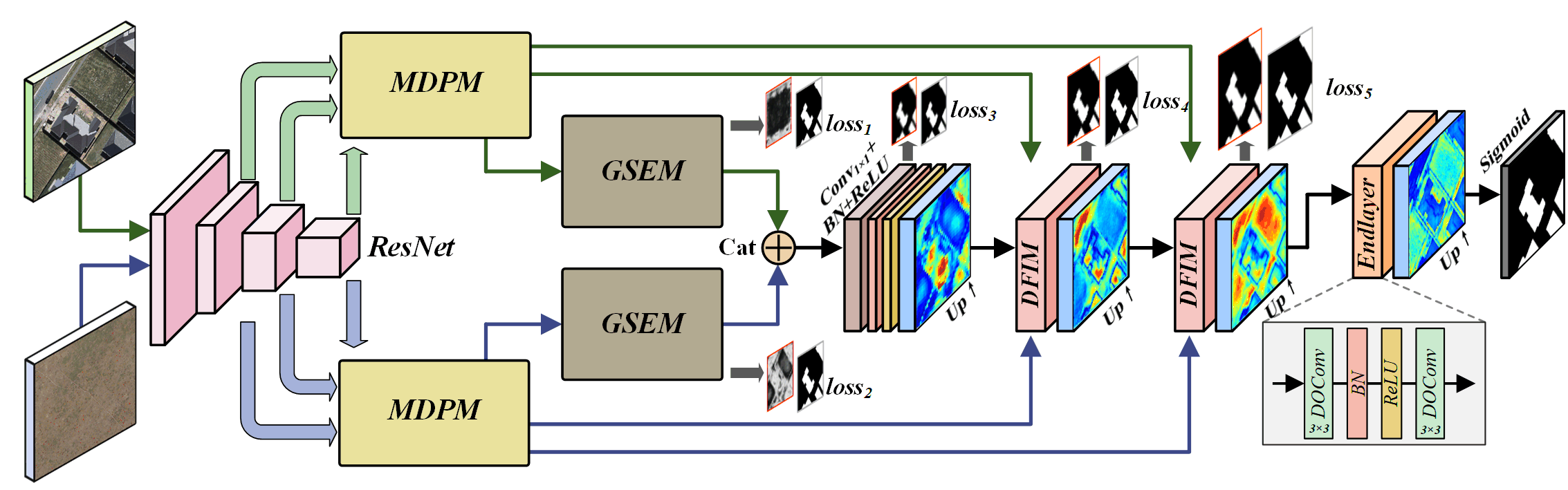}
		\end{tabular}
	\end{center}
	\caption
	{ \label{FIG:2}
		MFDS-Net master network diagram, $\text{Re}sNet_{34}$ as the backbone network completes feature extraction. Feature enhancement of feature information at different scales is performed by MDPM. GSEM accomplishes the integration of contextual feature information from a global viewpoint. DFIM highlights difference features. Enhancing the network's training procedure through the implementation of a deep supervision mechanism. }
\end{figure*}

Figure \ref{FIG:1} shows the experimental results of IFNet, ChangeFormer, and Ours on the LEVIR-CD \cite{20}, WHU-CD \cite{21}, and GZ-CD \cite{22} datasets. When facing complex situations such as dense and diverse targets, the performance of IFNet and ChangeFormer is not satisfactory. However, Ours shows more ideal results in edge processing and change target recognition.

In this paper we have made the following main contributions:

(1) We propose a MFDS-Net in Figure \ref{FIG:2}. MDPM is introduced to enrich the texture and position information contained in the features and strengthen the attention to detail features.

(2) A GSEM is constructed to enhance the correlation between high-level semantic information from a global perspective. A DFIM is created in the change target reconstruction stage to fuse high-level semantic information with two different spatiotemporal feature information and enhance the focus on change targets.

(3)DO-Conv and deep supervision mechanisms are employed to enhance the training performance of the network, and the results show that MFDS-Net achieves better performance than mainstream methods on the LEVIR-CD, WHU-CD, and GZ-CD datasets.

\section{Related Work}

In this section, we provide a synopsis of the evolution of convolutional neural networks in remote sensing change detection, explore the attention mechanism, and conclude with a concise overview of the concept of a deep supervision mechanism.

\subsubsection{CNN and Remote Sensing Change Detection}

Deep learning methods, represented by Convolutional neural networks (CNN), learn differences in dual-time images through efficient automatic processing, resulting in clearer, more explicit results and more detailed descriptions of the edges of the difference target. And CNN has been widely used in the field of computer vision \cite{xu2019lecture2note, xu2022arbitrary}.

FC-EF \cite{23} is based on U-Net \cite{24}, which follows the classical "encoder-decoder" structure in segmentation. It is worth noting that in FC-EF, two different spatiotemporal input features are fused in the form of feature splicing before entering the network. This approach destroys the information of the respective features in the two different spatio-temporal images, so that much of the detailed information becomes blurred.

FC-Siam-Conc \cite{23} uses separate encoders for feature extraction of different spatio-temporal images. This processing maximises the integrity of the feature information in the two different spatio-temporal images. It has also become a classical processing method in the field of CD. FC-Siam-Di \cite{23} subtracts the two features used for the skip connection and passes them to the decoder for feature fusion. This processing idea became another important processing idea in the field of CD. Since the convolution operation can only capture the correlation between local features, the results shown by these two methods are not ideal.

The ideas of UNet++ \cite{25} have also been applied in the field of CD. SNUNet \cite{26} uses the idea of dense connectivity in UNet++ to incorporate the texture information contained in low-level semantic information into high-level semantic information, enriching the attention and learning of the deep network for detailed information. In addition, an ECAM is built into SNUNet to interact and fuse feature information across multiple outputs. Nevertheless, there remains scope for enhancing the processing of intricate scenes and detailed information. HMLNet \cite{27} proposes a metric-based learning network, which takes the multi-scale feature information obtained by the backbone network and passes it into a feature pyramid for metric learning. Subsequently, it bolsters the capability to concentrate on feature information through the dual attention module.

MSCANet \cite{28} uses the Transformer to create an MSCA module to process feature information at different scales, enhancing the focus on global information. The deep supervision mechanism is also applied in MSCANet, optimising the training process by calculating losses on the output features at different scales. In BIT \cite{29}, the feature extraction of the dual spatio-temporal images begins with the utilization of ResNet \cite{30} as the backbone network; the feature information is transformed into semantic labels and fed into the Transformer encoder to reinterpret the label information from a global viewpoint. Subsequently, the Transformer decoder reshapes the labels into pixel features. Finally, the difference features are output after being processed by the convolutional layer.

Although CNNs have yielded numerous results in the field of CD, our experiments with most of the mainstream networks revealed that they are still somewhat deficient in their handling of complex targets. Therefore, we propose MFDS-Net on top of the CNN architecture.

\subsubsection{Attention Mechanism} 

Although VGG, FCN\cite{31}, U-Net and other single convolutional neural networks (CNNs) have achieved good results when migrating to the CD domain. The performance of the network cannot break through the bottleneck because it is limited by the convolutional kernel, which can only establish associations for local feature information and cannot take into account remote dependencies. Attention mechanisms such as self attention, channel attention \cite{32}, SK attention \cite{33} have emerged in order to capture richer contextual information and retain more details and location information when reconstructing feature information. By embedding the attention mechanism, CNNs enhance the emphasis on feature information, resulting in finer prediction results and more accurate target locations.

DTCDSCN \cite{34} uses SE-ResNet as backbone network to perform feature extraction, where the channel attention enhances the feature strength from a global perspective. In addition the DTCDSCN also introduces the dual attention module (DAM) in DTCDSCN for the extraction of disparity features.

Considering that the correlation with environmental noise is also calculated when associating global features, we utilize channel attention and Non-local block \cite{35} to construct GSEM in the bottleneck layer of MFDS-Net to enhance the correlation between contexts in the deep network, suppressing irrelevant channel information to the change target and enhancing the channels where the change target is located.

\subsubsection{Deep Supervision Mechanism} 

In CNNs, the desire to obtain richer feature information often requires the design of deeper network structures, and deeper networks may suffer from gradient disappearance problems. To cope with such problems, DSN\cite{36} proposes the idea of deep supervision. Specifically, throughout the training of the network, concomitant outputs are set at important processing and the loss between these concomitant outputs and the GT is calculated so that the hidden layer of the network can be supervised. In addition, the DSN demonstrates that the deep supervision mechanism works equally well for training large datasets.

As in IFNet, SNUNet, the same deep supervision mechanism is used in ADS-Net\cite{37}. a corresponding decoder is set up in ADS-Net for the different scales of features acquired by the encoder for differential feature extraction, and the results are added to the supervision mechanism to calculate the loss.

A depth-supervision mechanism is also used in MFDS-Net. Considering that CD involves two different images, we set up a deep supervision mechanism for the extraction process of both feature information, so as to optimise the network training process.

\section{Method}\label{sec3}
In this section, the overall structure of the network is first introduced. This is followed by an introduction to DO-Conv and a brief description of MDPM, GSEM and DFIM. Finally I will show the algorithmic flow of the network, as in Algorithm \ref{alg1}.

\subsection{Network Structure}

MFDS-Net(as in Figure \ref{FIG:2}) comprises four stages. Firstly, the feature extraction phase uses $\text{Re}sNet_{34}$ as backbone network to perform the feature extraction of the dual spatio-temporal images. We reconstructed $\text{Re}sNet_{34}$ by eliminating maximum pooling and average pooling, in order to retain richer information on weak features for contextual association in subsequent processing. We adopt DO-Conv \cite{38} to replace all the convolutions in the network in order to reduce the computational effort during training and to improve the training effect of the network.

Next, we feed feature information at different scales into the MDPM and concentrate the feature information through residual blocks constructed by dense convolution. The parallel multi-scale receptive field convolution blocks are used to enhance the correlation between feature information, in order to overcome the constraint on the field of view imposed by the fixed convolutional kernel. In order to maintain feature integrity and prevent the gradient disappearance problem, the entire module still uses the residual structure.

Subsequently, considering that high-level semantic information can enrich the contour information and location information of the whole target from a global viewpoint. At the bottleneck level, we form GSEM using channel attention and Non-local block to enhance the contextual association of the network with high-level semantic information. At this point, our dual spatiotemporal image feature extraction process is over. In this process, we have enhanced the feature information of different scales in order to create the most favorable data conditions for the next differential feature extraction.

In the final stage of the network, we begin by integrating high-level semantic information through feature stitching. Then, the convolutional downsampling and upsampling are inputted into DFIM to fuse the high-level semantic information with both types of low-level semantic information and perform differential feature extraction. Finally, the Endlayer processing yields the ultimate prediction results.

Regarding the deep supervision mechanism, since the input images are two different images at different times, it is possible that the disparity target is contained in both images. Therefore, we set concomitant outputs for the two different features from the bottleneck layer onwards. In the difference feature extraction stage, concomitant outputs are also set for feature information at different scales. Finally we calculate the losses for all the concomitant outputs and weight them to the total loss function.

\subsection{DO-Conv}

\begin{figure}
	\begin{center}
		\begin{tabular}{c}
			\includegraphics[height=5cm]{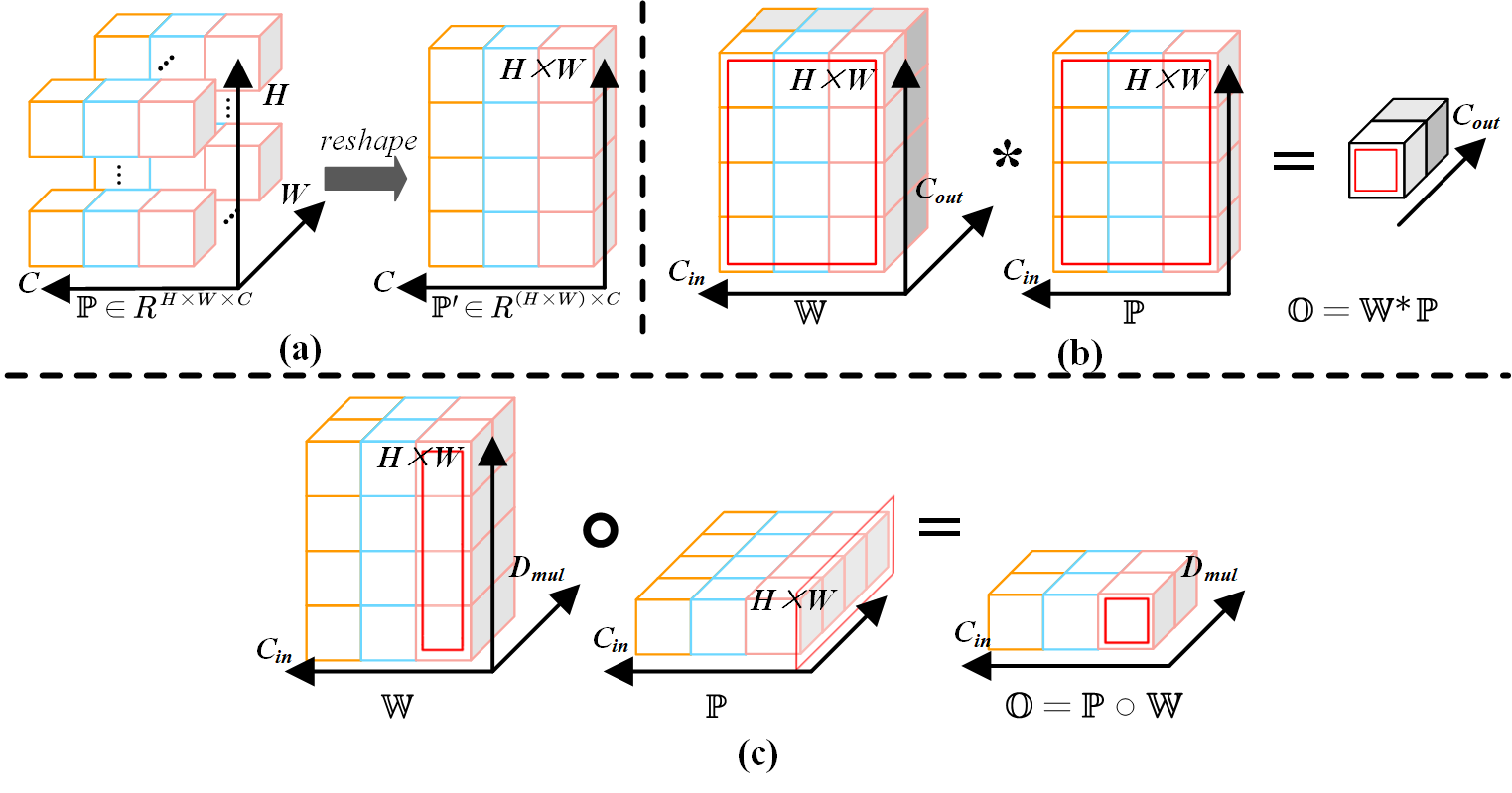}
		\end{tabular}
	\end{center}
	\caption
	{ \label{FIG:3}
		(a) shows the specific process of feature reconstruction. (b) shows the traditional convolution process. (c) shows the Depthwise convolution process. }
\end{figure}

As in Figure \ref{FIG:3}(a), 3D tensor $\mathbb{P}\in R^{H\times W\times C}$ can be reconstructed as 2D tensor $\mathbb{P}'\in R^{\left( H\times W \right) \times C}$ while maintaining the original data, where \textit{H}, \textit{W} is the space size and \textit{C} is the number of channels. Conventional convolution is often performed on local feature information in the form of a sliding window during image processing. As in Figure \ref{FIG:3}(b), we demonstrate the exact process of a single convolution operation. The area of the image covered by the convolution layer can be defined as \textit{patch}($\mathbb{P}\in R^{H\times W\times C_{in}}$), where $C_{in}$ is the number of input channels for the feature. The convolution layer can then be defined as 3D tensor($\mathbb{W}\in R^{C_{out}\times \left( H\times W \right) \times C_{in}}$), where $C_{out}$ is the number of channels of the output features. The result of this calculation is $\mathbb{O}_1$ with a channel count of $C_{out}$(as in Eq.~\ref{eq:1}).

\begin{figure}
	\begin{center}
		\begin{tabular}{c}
			\includegraphics[height=6cm]{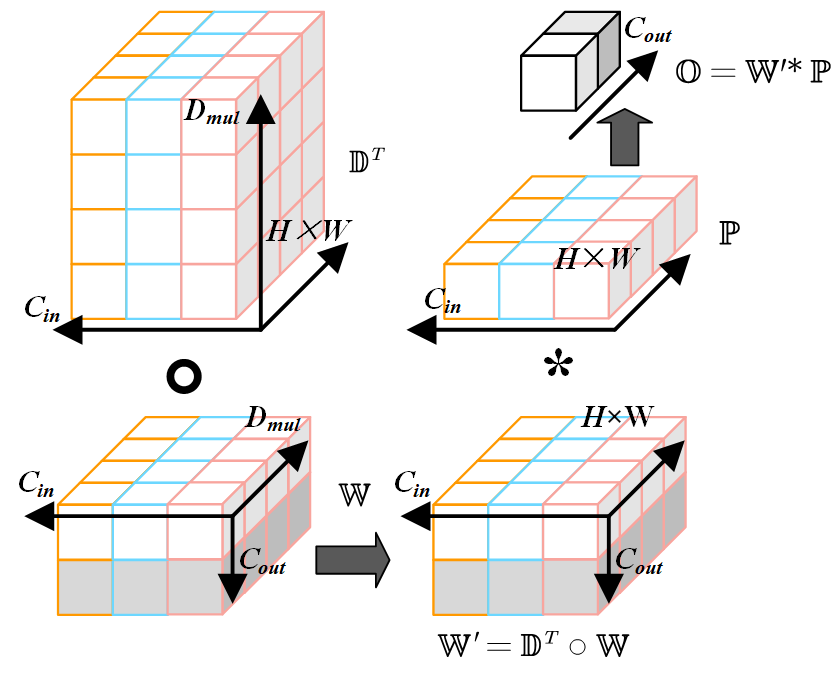}
		\end{tabular}
	\end{center}
	\caption
	{ \label{FIG:4}
		The DO-Conv process. }
\end{figure}

\begin{equation}
	\mathbb{O}_{1C_{out}}=\sum_{}^{\left( H\times W \right) \times C_{in}}{\mathbb{W}_{C_{out}i}\mathbb{P}_i}
	\label{eq:1} 
\end{equation}

In Figure \ref{FIG:3}(b), the final result is obtained by summing the multiplication calculations at the corresponding positions in the regions indicated by the red boxes. Where, $C_{in}=3$ is labelled in three different colours. $C_{out}=2$, is distinguished in grey.

The Depthwise convolution \cite{39} idea is shown in Figure \ref{FIG:3}(c), where the convolution layer $\mathbb{W}\in R^{\left( H\times W \right) \times D_{mul}\times C_{in}}$ (where $D_{mul}$ is the depth multiplier) is computed with $\mathbb{P}\in R^{H\times W\times C_{in}}$ to obtain $\mathbb{O}_2$(as in Eq.~\ref{eq:2}).The corresponding result is obtained by multiplying and adding the positions shown in red in Fig.~\ref{FIG:3}(c).

\begin{equation}
	\mathbb{O}_{2D_{mul}C_{in}}=\sum_{}^{\left( H\times W \right)}{\mathbb{W}_{iD_{mul}C_{in}}\mathbb{P}_{iC_{in}}}
	\label{eq:2} 
\end{equation}

DO-Conv merges traditional convolutional computation with depthwise convolutional computation, as in Figure \ref{FIG:4}. Firstly, DO-Conv multiplies the Depthwise convolution kernel operator with the traditional convolution kernel operator to obtain the result $\mathbb{W}'$. Finally, $\mathbb{W}'$ is used to perform the ordinary convolution operation with $\mathbb{P}$ to obtain the final result $\mathbb{O}$(as in Eq.~\ref{eq:3}).

\begin{equation}
	\mathbb{O}=\left( \mathbb{D},\mathbb{W} \right) \circledast \mathbb{P}=\left( \mathbb{D}^T\circ \mathbb{W} \right) \ast \mathbb{P}
	\label{eq:3} 
\end{equation}

$\mathbb{D}\in R^{\left( H\times W \right) \times D_{mul}\times C_{in}}$ is the convolution operator for Depthwise convolution and $\mathbb{W}\in R^{C_{out}\times D_{mul}\times C_{in}}$ is the convolution operator for conventional convolution. $\mathbb{D}^T\in R^{D_{mul}\times \left( H\times W \right) \times C_{in}}$ is the transpose of $\mathbb{D}$.

In MFDS-Net, we adopt DO-Conv to replace all traditional convolution operations, aiming to enhance the network's training process. Another important purpose of using DO-Conv is to reduce the computational effort of the network.

\subsection{Multi-scale Detail Preservation Module}

\begin{figure*}
	\begin{center}
		\begin{tabular}{c}
			\includegraphics[height=5cm]{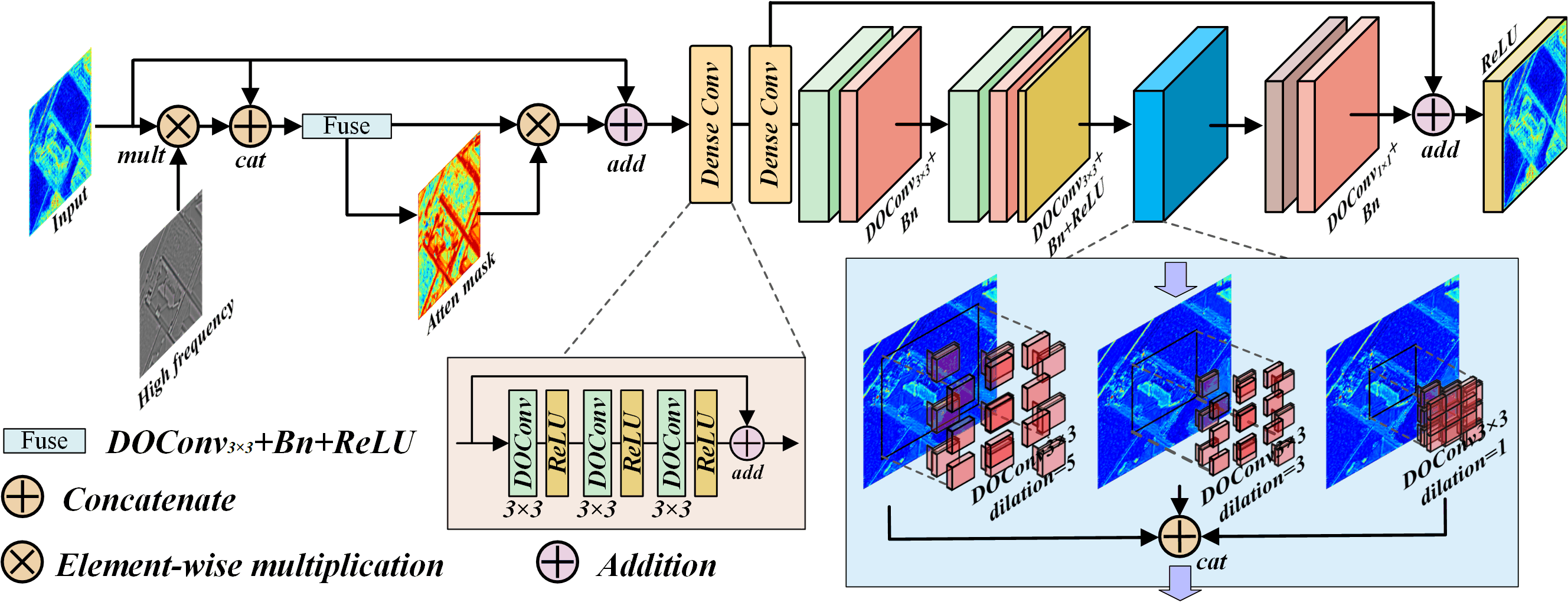}
		\end{tabular}
	\end{center}
	\caption
	{ \label{FIG:5}
		Multi-scale Detail Preservation Module(MDPM). }
\end{figure*}
As shown in Figure \ref{FIG:5}, we have developed MDPB and applied it after the backbone network, enhancing the acquired spatiotemporal image features using it. To robustly preserve edge information, we utilize the Laplacian pyramid method to retain high-frequency details (such as edges and contours) in the image. Considering the sensitivity of the Laplacian operator to noise, we first apply Gaussian smoothing to the image, followed by the Laplacian operation, to obtain the high-frequency feature representation, denoted as $ f_h\in\mathbb{R}^{H\times W\times1}$. Then, we perform element-wise multiplication between $f_h$ and the output $f\in\mathbb{R}^{H\times W\times C}$ of the backbone network at any scale, yielding the edge-enhanced feature $f_e{\in\mathbb{R}}^{H\times W\times C}$. 
\begin{equation}
	f_e=f_h\otimes f
\end{equation}

\begin{equation}
	f_{fuse}=Conv[f_e\otimes f]
	\label{eqnew1}
\end{equation}
Afterwards, we fuse $f$ and $f_e$ through convolution to obtain $f_{fuse}$ as shown in Eq.\ref{eqnew1}, where $\otimes$ represents element-wise multiplication operation, and $\left[ \cdot \right]$ represents concatenation operation along the channel dimension. Considering that high-frequency information may contain potential noise and redundant details, we introduce an attention mask $A_m$ to focus the model's attention on critical regions while suppressing background noise and redundant information. The definition of $A_m$ is as shown in Eq.\ref{eqam}. Finally, the edge-enhanced feature can be represented by the following formula:
\begin{equation}
	A_m=\sigma(Conv(f_{fuse}))
\end{equation}

\begin{equation}
	{\hat{f}}_i=g\times f_{fuse}\otimes A_m+f_i
	\label{eqam}
\end{equation}
Where $\sigma$ represents the Sigmoid function, $g$ represents learnable parameters that enable the model to automatically learn appropriate edge-enhanced features. Refer to the work in \cite{mega}, we subsequently pass the features through CBAM for recalibration. This step helps capture feature correlations between boundaries and background regions, as shown in the formula.

\begin{equation}
	X_i=CBAM({\hat{f}}_i)
\end{equation}

Afterwards, we constructed densely connected convolutional blocks predominantly using residual structures to concentrate feature information. Considering that traditional small convolutional kernels are limited by the size of the kernel and cannot establish long-range dependencies between pixels, while using large convolutional kernels would increase computational complexity during training. Therefore, we aimed to expand the receptive field of convolutions without increasing computational cost. Specifically, we implemented a three-way parallel processing approach, assigning different dilation factors to each convolution path to obtain receptive fields of different scales. This approach allows the convolutional kernels to span different distances between pixels at different dilation factors, thereby capturing image information at various scales. It enables effective modeling of correlations between pixels even at distant positions.

As in Eq.~\ref{eq:4}, the feature $X\in \mathbb{R}^{H\times W\times C}$ first passes through the dense convolution layer to complete the feature set and obtain the result $X'\in \mathbb{R}^{H\times W\times C}$, where $f_D\left( \cdot \right)$ is the dense convolution block.

\begin{equation}
	X'=f_{D_i}\left( f_{D_{i-1}}\left( X \right) \right) ,i=2
	\label{eq:4} 
\end{equation}

Subsequently, as in Eq.~\ref{eq:5}, the multi-scale perceptual field convolution completes the parallel processing of $X'$. The feature fusion is completed by dimensional splicing to obtain $X_F$, where $f_{1\times 1}\left( \cdot \right)$ is the 1$\times$1 convolution and $f_r\left( \cdot \right)$  is the different scale perceptual field convolution.

\begin{equation}
	X_F=f_{1\times 1}\left( f_{r=5}\left( X' \right) +f_{r=3}\left( X' \right) +f_{r=1}\left( X' \right) \right) 
	\label{eq:5} 
\end{equation}

Finally, the enhanced features are weighted onto $X'$ to obtain the final result $X_O$, as in Eq.~\ref{eq:6}, $\alpha$ where is $\text{Re}LU$.

\begin{equation}
	X_O=\alpha \left( X_F+X' \right)  
	\label{eq:6} 
\end{equation}

\begin{figure}
	\begin{center}
		\begin{tabular}{c}
			\includegraphics[height=4.2cm]{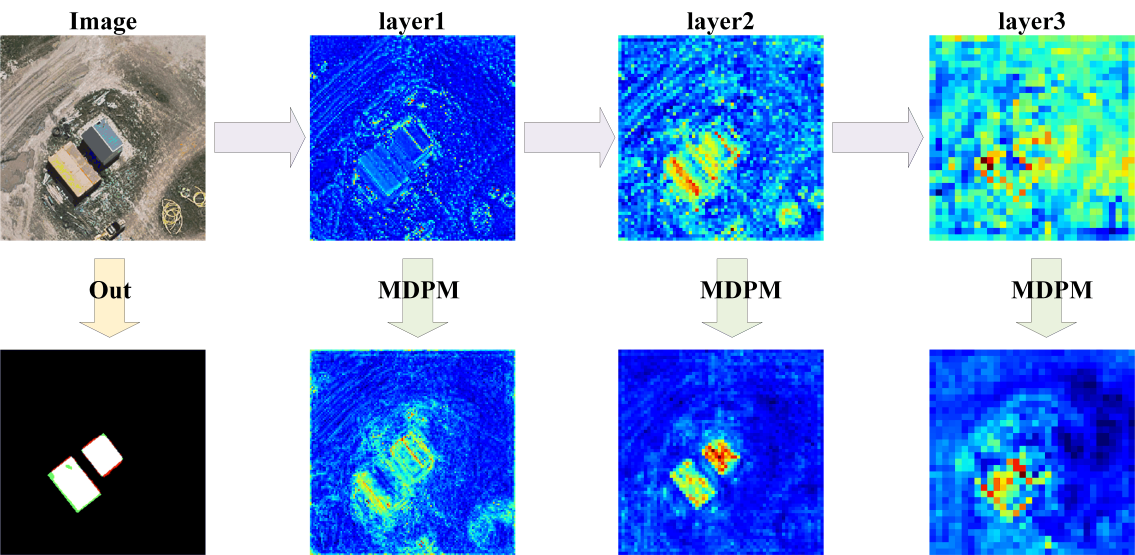}
		\end{tabular}
	\end{center}
	\caption
	{ \label{FIG:6}
		Heatmap obtained after MDPM processing. $layer_i$ represent the three different levels of feature information obtained by $\text{Re}sNet_{34}$ respectively. }
\end{figure}

As in Figure \ref{FIG:6}, the features are significantly enhanced after MDPM processing, where the weight of the target feature information is significantly increased, especially the edge information of the target. In addition, the high-level semantic information captured by the backbone network was fuzzy and could not locate the target location, whereas after MDPM processing, the target location was clearly marked.

\subsection{Global Semantic Enhancement Module}

\begin{figure}
	\begin{center}
		\begin{tabular}{c}
			\includegraphics[height=3.4cm]{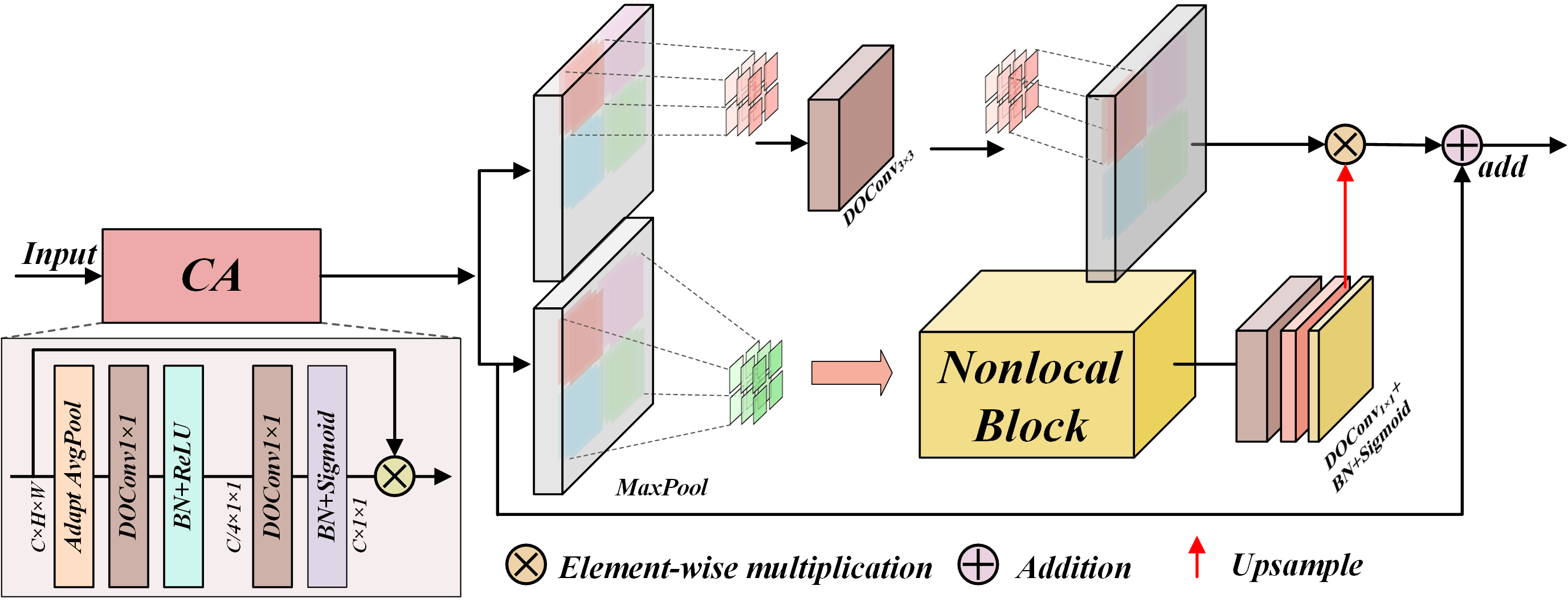}
		\end{tabular}
	\end{center}
	\caption
	{ \label{FIG:7}
		Global Semantic Enhancement Module(GSEM). }
\end{figure}

As in Figure \ref{FIG:7}, the macro feature information contained in the high-level semantic information plays an important role in the reconstruction of disparate features due to its importance. Therefore, we form the Global Semantic Enhancement Module (GSEM) in the bottleneck layer using channel attention (CA) and Non-local block to enhance the high-level semantic information.

In change detection tasks, images may exhibit changes due to different scenes and environmental factors, such as variations in lighting and seasons. Changes in lighting and seasons can alter the appearance and texture of objects in images, but their higher-level semantic information often remains unchanged, such as the shape, position, and size of objects. Therefore, we aim to enhance higher-level semantic information to focus the model on the core features of the objects, thus improving the robustness and accuracy of change detection. Specifically, in the bottleneck layer, we first adjust the correlation between feature channels using Channel Attention Block to optimize the representation of global semantic information, aiding the model in selecting and adjusting feature representations. Subsequently, we construct a Semantic Context Module (SCM) through Non-local Block, further enhancing the global connectivity of features, enabling the model to better understand the target structures and semantic information in the images.

In the Channel Attention Block, the input feature $X\in \mathbb{R}^{H\times W\times C}$ is compressed to $Z\in \mathbb{R}^{1\times 1\times C}$, as shown in Eq.~\ref{eq:7}, by global averaging pooling.

\begin{equation}
	Z=\frac{1}{H\times W}\sum_{i=1}^H{\sum_{j=1}^W{X_{\left( i,j \right)}}}
	\label{eq:7} 
\end{equation}

After two fully connected layers with a convolution kernel size of 1$\times$1 to establish channel correlation, $Sigmoid$ is used to normalize the channel weights to obtain $Z_{norm}\in \mathbb{R}^{1\times 1\times C}$, as in Eq.~\ref{eq:8}, where $\sigma$ is $Sigmoid$. Finally, each channel weight of $Z_{norm}$ is multiplied with each channel of the original feature $X$ to obtain the final result $X_O$, as in Eq.~\ref{eq:9}.

\begin{equation}
	Z_{norm}=\sigma (f_{1\times 1}\left( \alpha f_{1\times 1}\left( Z \right) \right) )
	\label{eq:8} 
\end{equation}
\begin{equation}
	X^\prime=Z_{norm}\otimes X
	\label{eq:9} 
\end{equation}

In SCM, we establish two branches, as illustrated in Figure \ref{FIG:7}. In the upper branch, the feature $ X^\prime \in \mathbb{R}^{H\times W\times C} $ is initially divided into $ k\times k $ patches of size $ \frac{W}{k}\times\frac{H}{k} $. We apply convolution on each patch to capture semantic information locally. Subsequently, each patch is reorganized back into the original order, forming $ X^{\prime\prime}\in\mathbb{R}^{H\times W\times C} $. 

In the lower branch, the input feature undergoes pooling to obtain features of size $ k\times k $, denoted as $ W $. Then, through the Non-local Block operation, contextual information within the local region is analyzed to acquire $ W^\prime $. Each pixel in $ W^\prime $ represents the corresponding patch in the upper branch. Following this, interpolation is applied to $ W^\prime $ to perform upsampling, yielding $ W^{\prime\prime} $. After upsampling, $ W^{\prime\prime} $ is element-wise multiplied with $ X^{\prime\prime} $, and then multiplied by a learnable parameter $ \gamma $ to enable the network to select more effective features, as depicted in Eq.\ref{eq:gasb-1}.

\begin{equation}
	X_e=\gamma\times W^{\prime\prime}\otimes X^{\prime\prime}
	\label{eq:gasb-1} 
\end{equation}

\begin{equation}
	X_O=f_{1\times1}(Cat\left(X_e^i,...,X_e^j\right))
	\label{eq:gasb-2} 
\end{equation}

In GSAB, we establish four sets of parallel SCM, each with a different setting for $ k $. At the end of each module, we concatenate the outputs of the four sets of SCM and pass them through a convolution operation to generate the final output $ X_O $, as shown in Eq.\ref{eq:gasb-2}. Here, $ i $ and $ j $ represent the indices of different SCM outputs, and $ f_{1\times1} $ denotes a $ 1\times1 $ convolution.

\begin{figure}
	\begin{center}
		\begin{tabular}{c}
			\includegraphics[height=5cm]{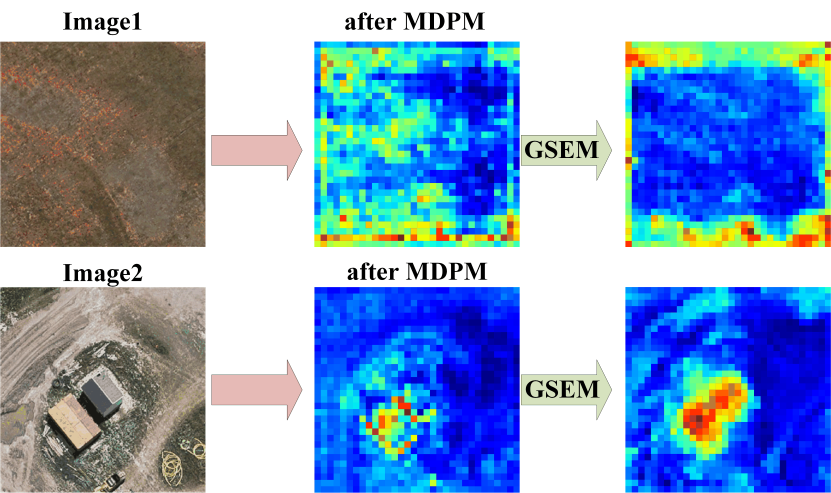}
		\end{tabular}
	\end{center}
	\caption
	{ \label{FIG:8}
		Heatmap obtained after GSEM treatment. }
\end{figure}

As in Figure \ref{FIG:8}, we show the results obtained by GSEM for the processing of high-level semantic information. After the GSEM processing, the relatively weak information in the target features is enhanced and the target profile information is better described in the high-level semantic information due to the establishment of remote dependencies.

\subsection{Differential Feature Integration Module}\label{subsec35}

\begin{figure}
	\begin{center}
		\begin{tabular}{c}
			\includegraphics[height=5.5cm]{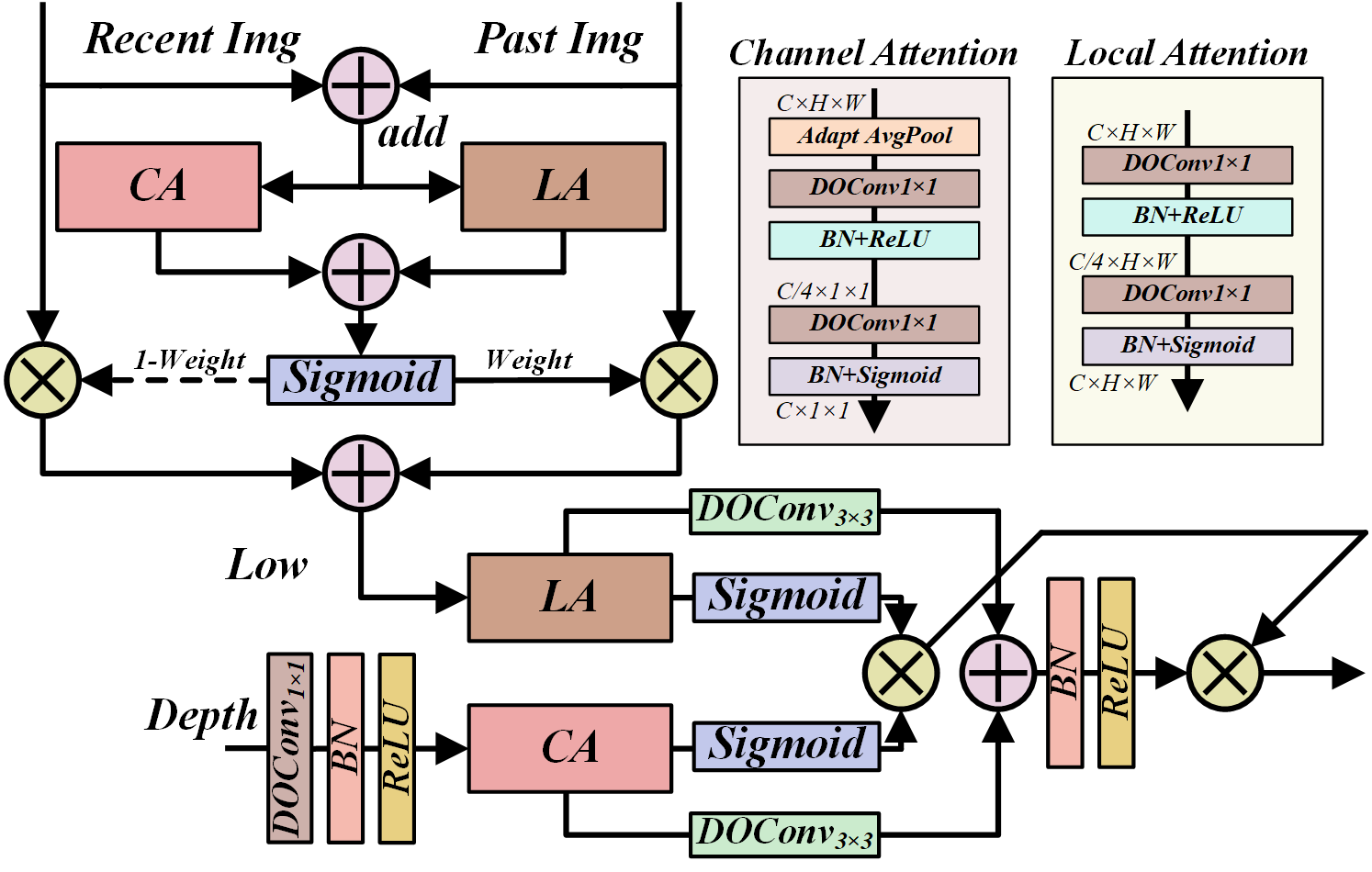}
		\end{tabular}
	\end{center}
	\caption
	{ \label{FIG:9}
		Differential Feature Integration Module(DFIM). }
\end{figure}

As in Figure \ref{FIG:9}, in order to complete the extraction of difference features, we created DFIM to receive the fusion results of two types of features from GSEM, and also receive shallow features of two different images. DFIM completes the fusion of three different dimensional features and completes the extraction of difference features.

DFIM is divided into two stages. In the first stage, we use AFF \cite{40} to complete the fusion of the two shallow features. In AFF, the fusion of the two features is completed by summing the values (as in Eq.~\ref{eq:12}) fed into the parallel CA and LA to complete the feature focus in terms of both channel weights and pixel weights. After the attention mechanism, the features are again summed and normalised by $Sigmoid$ to obtain the weights $W$, as in Eq.~\ref{eq:13}. Finally, the weights are assigned by matrix multiplication and summed to obtain the result $I_L$, as in Eq.~\ref{eq:14}. Where $f_C$ represents CA and $f_L$ represents LA.
\begin{equation}
	I_F=I_R+I_P
	\label{eq:12} 
\end{equation}
\begin{equation}
	W=\sigma \left[ f_C\left( I_F \right) +f_L\left( I_F \right) \right] 
	\label{eq:13} 
\end{equation}
\begin{equation}
	I_L=W\otimes I_P+\left( 1-W \right) \otimes I_R
	\label{eq:14} 
\end{equation}

In the second stage, considering that depth features contain more macroscopic feature information and their channel weights are of significant value, we use CA for depth features, while shallow features are richer in location and contour information of disparity targets, so we use LA for enhancement from a pixel perspective. Subsequently, we multiply (as in Eq.~\ref{eq:15}) and add (as in Eq.~\ref{eq:16}) the attended shallow feature $I_L$ and the deep feature $I_D$, respectively. Finally, $I_{add}$ is optimized by $BN$ and $\text{Re}LU$ and multiplied with $I_{mul}$ to obtain the final result $I_{out}$, as in Eq.~\ref{eq:17}.
\begin{equation}
	I_{mul}=\sigma f_L\left( I_L \right) \otimes \sigma f_C\left( I_D \right)
	\label{eq:15} 
\end{equation}
\begin{equation}
	I_{add}=f_L\left( f_{3\times 3}\left( I_L \right) \right) +f_C\left( f_{3\times 3}\left( I_D \right) \right)
	\label{eq:16} 
\end{equation}
\begin{equation}
	I_{out}=\alpha I_{add}\otimes I_{mul}
	\label{eq:17} 
\end{equation}

\begin{figure}
	\begin{center}
		\begin{tabular}{c}
			\includegraphics[height=6cm]{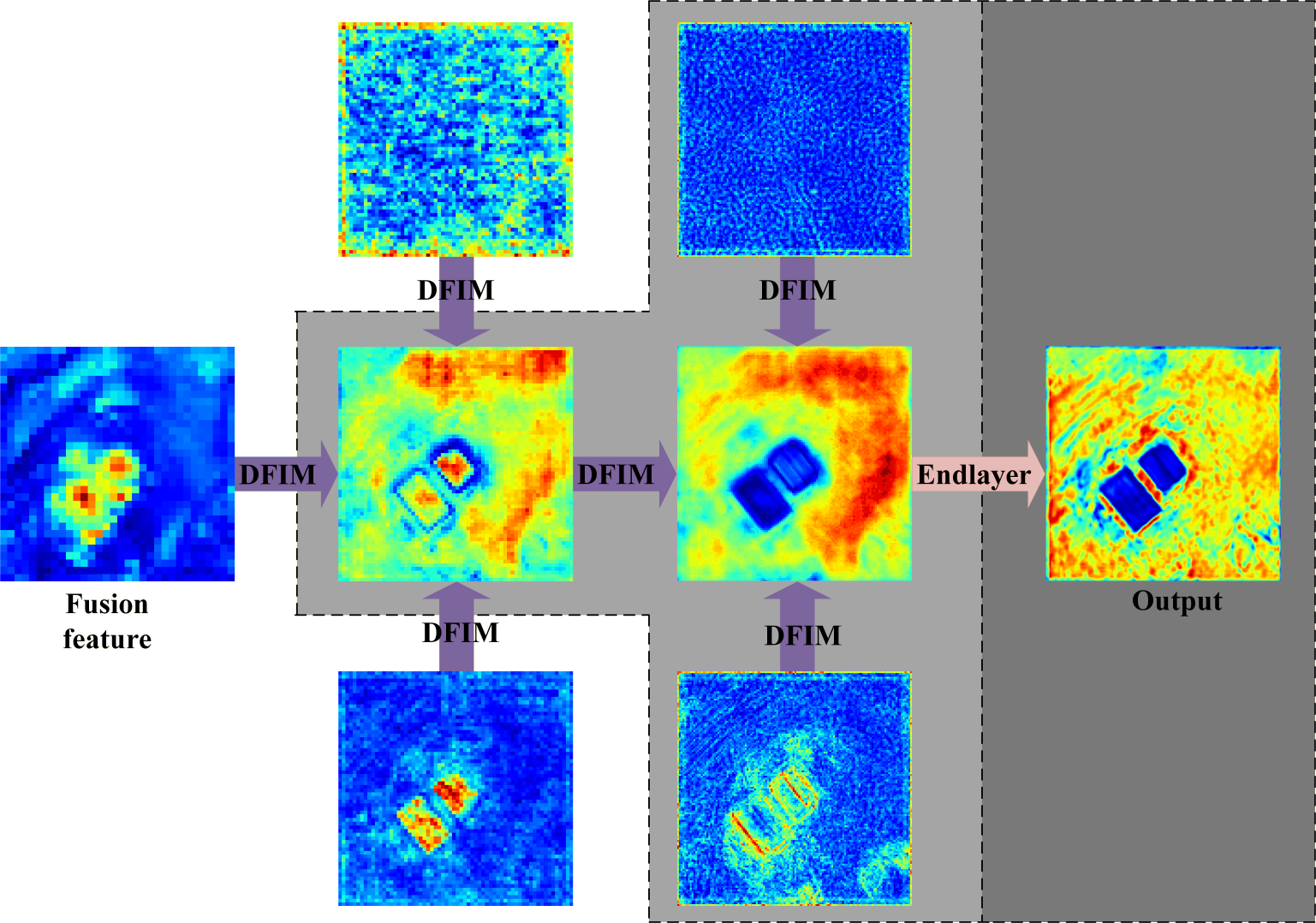}
		\end{tabular}
	\end{center}
	\caption
	{ \label{FIG:10}
		Heatmap obtained after DFIM treatment. }
\end{figure}

As in Figure \ref{FIG:10}, DFIM shows excellent performance throughout the processing, with its ability to effectively use the weight of the target location in the high-level semantic information for macroscopic description of the disparity features, using two different spatio-temporal feature information for localisation and edge description of disparity features.

\begin{algorithm}[H]
	\caption{MFDS-Net}\label{alg:alg1}
	\begin{algorithmic}
		\State 
		\State \textbf{Input:} $\text{Im}age_1=X_i\ i\in \left[ 1,n \right] $
		\State ~~~~~~~~~$\text{Im}age_2=Y_i\ i\in \left[ 1,n \right]$
		\State ~~~~~~~~~$GT=G_i\ i\in \left[ 1,n \right] $
		\State \textbf{Output:} $Segmentation\ Map\ S$
		\While{$not~converge$}
		\State Push(zip($X_i,Y_i,G_i$)) in
		\State $list_1=\text{Re}sNet_{34}\left( X_i \right)$
		\State $list_2=\text{Re}sNet_{34}\left( Y_i \right)$
		\State $mlist_1,mlist_2=MDPM\left( list_1,list_2 \right)$
		\State $out_1,out_2=GSEM\left( mlist_1\left[ 2 \right] ,mlist_2\left[ 2 \right] \right) $
		\State $Loss_1=loss\left( out_{\left( 1,i \right)},G_i \right)$
		\State $Loss_2=loss\left( out_{\left( 2,i \right)},G_i \right)$
		\State $Out=out_1+out_2$
		\State $Loss_3=loss\left( Out_i,G_i \right)$
		\State $Out_i=DFIM\left( Out_i\uparrow ,mlist_1\left[ 1 \right] ,mlist_2\left[ 1 \right] \right)$
		\State $Loss_4=loss\left( Out_i,G_i \right)$
		\State $Out_i=DFIM\left( Out_i\uparrow ,mlist_1\left[ 0 \right] ,mlist_2\left[ 0 \right] \right)$
		\State $Loss_5=loss\left( Out_i,G_i \right)$
		\State $S_i=Endlayer\left( Out_i \right) \uparrow $
		\State $Loss=loss\left( S_i,G_i \right)$
		\State $Loss_a=\theta \left( Loss_1+Loss_2 \right)$
		\State $Loss+=Loss_a +\varphi \left( Loss_3+Loss_4+Loss_5 \right)$
		\EndWhile
	\end{algorithmic}
	\label{alg1}
\end{algorithm}

\section{Experiment}

In this section, we first introduce the experimental parameters, datasets, and then briefly describe the comparative experiments and ablation experiments.

\subsection{Introduction to experimental parameters}

The entire training process for MFDS-Net is performed in a Linux environment, and multiple NVIDIA TITAN RTX graphics cards are used to provide the computing power for MFDS-Net. The pytorch version in the experiment is 1.4.0 and the python version is 3.8.

In addition, the number of iterations for all methods participating in the experiment was set to 200 rounds. We save the optimal model after each round of training. In terms of learning rate, the initial learning rate of all the methods involved in the experiments are 0.001, and Adam is used as the optimiser for the whole training process.

The experiments used binary cross loss entropy as the loss function (as in Eq.~\ref{eq:18}). Since MFDS-Net uses a deeply supervised mechanism (as in Figure \ref{FIG:2} loss), the total loss function is a weighted sum of the concomitant losses, as in Eq.~\ref{eq:19}, where $Loss$ is the loss of the final result, $Loss_1$ and $Loss_2$ are the losses calculated for the high-level semantic information of the two different spatio-temporal image. After comparing the experimental results with the specific convergence of the training process, we set the value of $\theta$ to 0.2 and the value of $\varphi$ to 0.5.

\begin{equation}
	\begin{aligned}
		L_{CE\left( p,\hat{p} \right)}=&-\frac{1}{WH}\sum_{x=0}^{W-1}{\sum_{y=0}^{H-1}{p\left( x,y \right) \log \hat{p}\left( x,y \right) }} \\
		&{{ +\left( 1-p\left( x,y \right) \right) \log \left( 1-\hat{p}\left( x,y \right) \right)}}
	\end{aligned}
	\label{eq:18}
\end{equation}
\begin{equation}
	L=\theta \left( Loss_1+Loss_2 \right) +\varphi \left( Loss_3+Loss_4+Loss_5 \right) +Loss
	\label{eq:19} 
\end{equation}

We use five evaluation metrics to subjectively evaluate the performance of each method. The five metrics are F1 (as in Eq.~\ref{eq:20}), IoU , Precision, Recall , and OA . The calculation of the five metrics is performed in PyCharm.

\begin{equation}
	F1=\frac{2TP}{2TP+FP+FN}
	\label{eq:20} 
\end{equation}
\begin{equation}
	IoU=\frac{TP}{FP+FN+TP}
	\label{eq:21} 
\end{equation}
\begin{equation}
	Precision=\frac{TP}{TP+FP}
	\label{eq:22} 
\end{equation}
\begin{equation}
	\text{Re}call=\frac{TP}{TP+FN}
	\label{eq:23} 
\end{equation}
\begin{equation}
	OA=\frac{TP+TN}{TP+TN+FN+FP}
	\label{eq:24} 
\end{equation}

\subsection{Experimental datasets}

LEVIR-CD \cite{20} is a large public dataset within the domain of CD. in the field of CD, sourced from the Google Earth platform, boasting high resolution (0.5 meters/pixel). The LEVIR-CD dataset covers the changes in some areas of Texas, USA, between 5 and 14 years. The samples in LEVIR-CD not only include large buildings but also detection targets of different shapes and color spaces. We perform non-overlapping default cropping on the original images with a size of 1024 × 1024, resulting in training set (7120), test set (2048), and validation set (1024) all resized to 256 × 256. After cropping, the scale and the number of complex samples in the dataset are further increased, which poses higher challenges to the network's performance.

WHU-CD \cite{21} contains a pair of aerial image samples sourced from Christchurch, New Zealand, with a size of 32507×15354. WHU-CD not only has a higher resolution but also includes about 22,000 buildings and other detailed information, which also poses challenges to the performance of the network. After cropping and random sampling of WHU-CD, a training set of 6096 samples, a test set of samples 762 and a validation set of 762 samples, all resized to a size of 256 × 256.

The GZ-CD \cite{22} contains VHR images ranging in size from 1006 $\times$ 1168 to 4936 $\times$ 5224, showing changes in the Guangzhou area of China over a ten-year period. It is captured from Google Earth using BIGEMAP and has a high resolution (0.55 m/pixel). After non-overlapping cropping, the sample size is 256 $\times$ 256. A random sampling of the sample space yields a training set of 2834 samples, a test set of 325 samples, and a validation set of 400 samples.

\subsection{Comparative Experiment}\label{subsec43}

\subsubsection{Comparison methods}

In this section we briefly describe the nine methods involved in the comparison experiments.

\subsubsection{LEVIR-CD}

\begin{figure*}
	\begin{center}
		\begin{tabular}{c}
			\includegraphics[height=6.8cm]{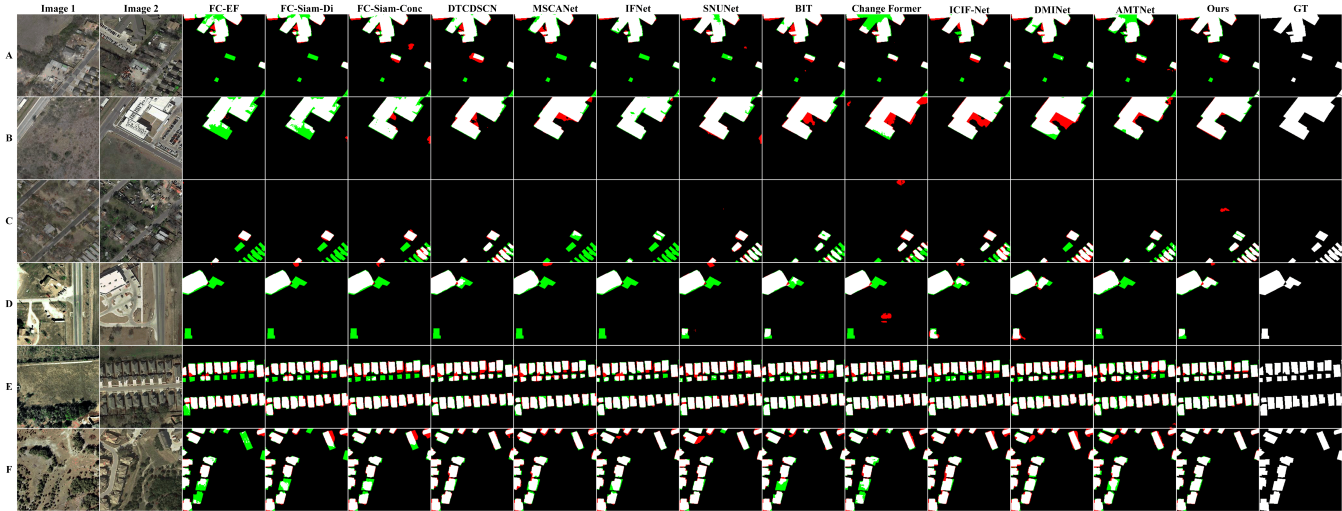}
		\end{tabular}
	\end{center}
	\caption
	{ \label{FIG:11}
		Graph of prediction results obtained by each method on the LEVIR-CD dataset.}
\end{figure*}

FC-EF\cite{23} is developed on the basis of U-Net. The data completes the fusion process before entering the network. Although FC-EF has achieved some success, FC-EF is significantly less capable of detecting targets in complex environments because the fusion process corrupts some complex feature information.

FC-Siam-Conc\cite{23} takes into account the problems with FC-EF and uses separate encoders for feature extraction from different spatio-temporal images to ensure the integrity of the feature information. This is still done in the form of feature summation in the subsequent processing. 
FC-Siam-Di\cite{23} subtracts the two features used for the skip connection and passes them to the decoder for feature fusion. This processing idea has become another important processing idea in the field of change detection.

DTCDSCN\cite{34} utilizes channel attention to boost the strength of feature information from a global viewpoint, and its extraction of feature information and sharing of weight information through twin neural networks has achieved good results.

IFNet\cite{15} uses the channel attention module and the spatial attention module to make the performance greatly improved. In addition, in IFNet, a deep supervision mechanism is used.

SNUNet\cite{26} uses the same architecture as UNet++ to prevent the semantic divide problem by fusing semantic information of different depths. the use of Ensemble Channel Attention Module (ECAM) enhances the representation of disparate features.

MSCANet\cite{28} is a discrepancy feature extraction model based on Transformer, combining convolutional neural networks with Transformer and using multi-scale processing to refine the feature information.

BIT\cite{29} employs ResNet as backbone network for feature extraction, and Transformer is used to provide global attention to deep information. Although BIT has achieved excellent results, there is instability in its processing of small targets as Transformer only focuses on deep feature information, and no attention is paid to shallow features.

Change Former\cite{18} differs from BIT in that it directly employs Transformer as backbone network for feature extraction, which is relatively robust in its ability to focus on small targets. Although better results are achieved, the number of parameters and floating point calculations are increased.

ICIF-Net\cite{ICIF} is a CNN-Transformer-based approach that effectively addresses the limitations of CNNs in modeling long-range dependencies by leveraging the strengths of both CNNs and Transformers. The network introduces mechanisms for local-global feature interaction and inter-scale fusion, where the Conv Attention module facilitates the extraction of local and global features at the same spatial resolution through interactive processing. Additionally, ICIF-Net adopts two attention-based inter-scale fusion schemes. Finally, the integrated features are fed into a conventional change prediction head to generate the final change detection results.

DMINet\cite{dminet} effectively addresses the foreground-background class imbalance and interference issues in remote sensing image change detection by combining self-attention and cross-attention mechanisms into the JointAtt module. Its key design includes acquiring difference features and multi-level difference aggregation, achieving superior performance compared to other methods while maintaining a simplified architecture, thus providing a reliable solution for remote sensing image change detection tasks.

AMTNet\cite{amtnet} is a CNN-Transformer-based method that leverages CNN for extracting multi-scale features. In the later stages of the network, attention mechanisms and Transformers are used to model contextual information within the images.

Figure \ref{FIG:11} shows the comparison experiments on the LEVIR-CD dataset. The results have been visually overlaid with GT in order to highlight detailed information. In this case, unrecognised regions are labelled in green, while misrecognised regions are labelled in red.

In sample A, the results obtained by FC-EF, SNUNet, ICIF-Net, Change Former, DMINet and AMTNet clearly performed poorly in terms of large target continuity. On the other hand, Ours demonstrates superior performance in maintaining the continuity of target features. In sample B, the buildings are affected by light and there are large shadows around them. However, networks such as BIT, Change Former, ICIF-Net, and DMINet do not provide satisfactory detailing of shadows. Ours, on the other hand, obtained the highest similarity to GT, and despite the room of improving edge detail, it still maintains a leading position in overall recognition.

\begin{figure}
	\begin{center}
		\begin{tabular}{c}
			\includegraphics[height=6cm]{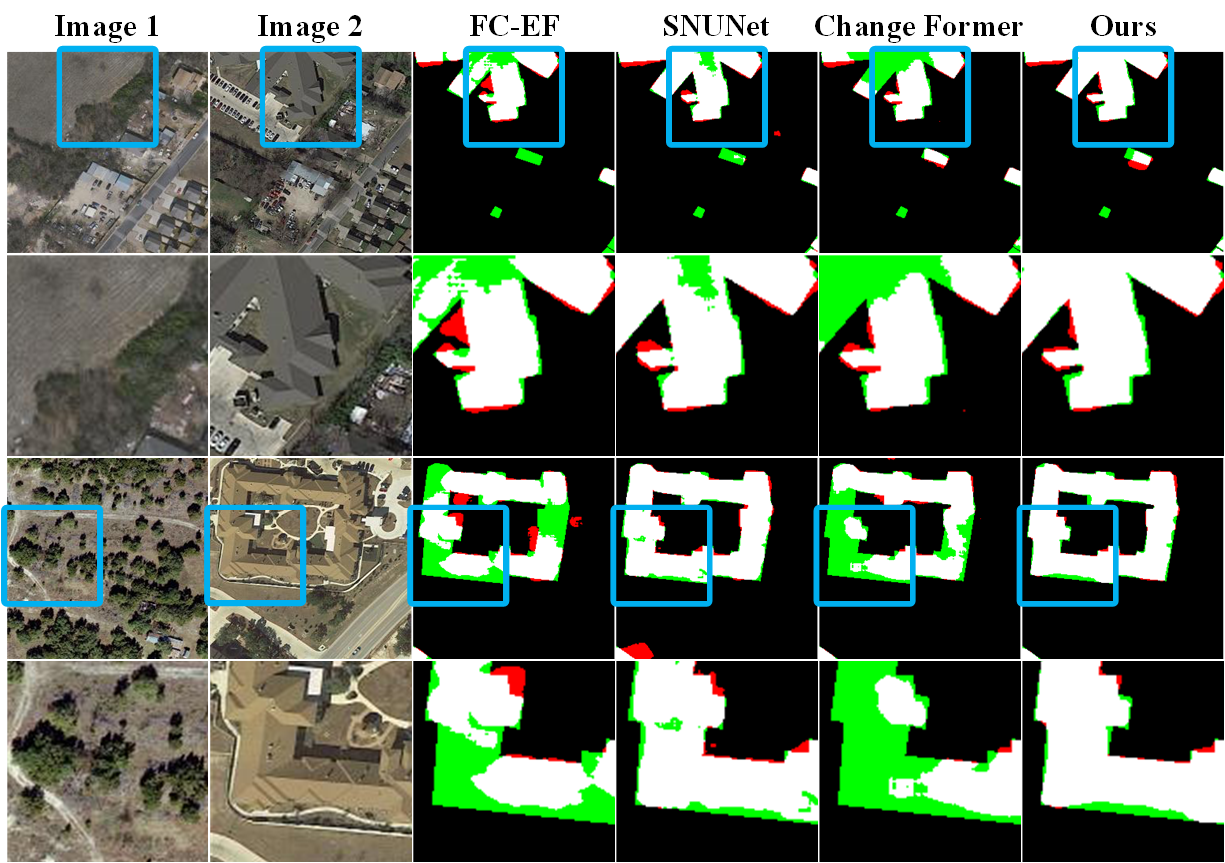}
		\end{tabular}
	\end{center}
	\caption
	{ \label{FIG:12}
		Detailed comparison chart of large targets.}
\end{figure}

In Figure \ref{FIG:12}, we compare Ours with three other networks, FC-EF, SNUNet and Change Former, for local zooming. As mentioned earlier, in the first sample, Ours exhibits better performance in terms of local feature continuity. In another sample, FC-EF, SNUNet, and Change Former still do not perform as well as Ours in terms of local feature continuity. 

\begin{figure}
	\begin{center}
		\begin{tabular}{c}
			\includegraphics[height=6cm]{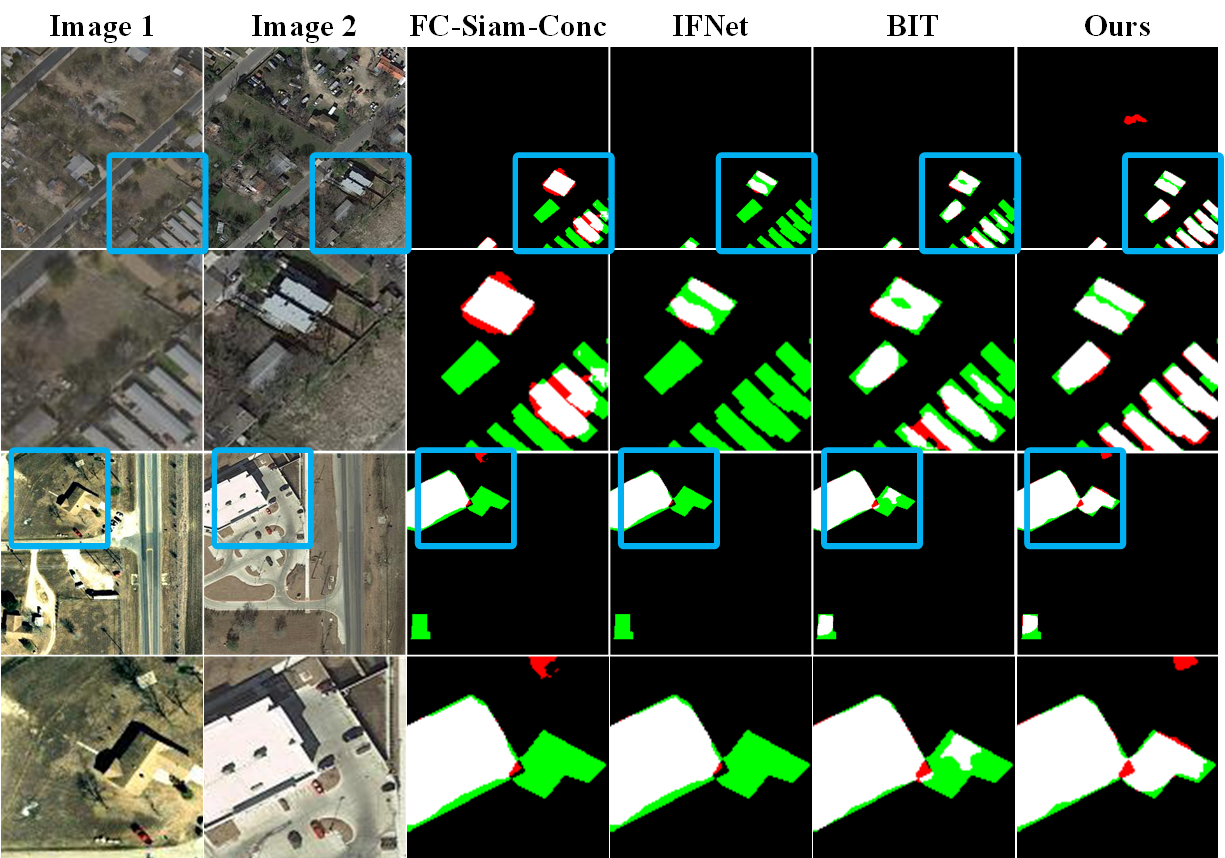}
		\end{tabular}
	\end{center}
	\caption
	{ \label{FIG:13}
		Comparison of partial enlargements of the detailing of complex variation samples.}
\end{figure}

Since the targets in the samples may exist in two different images at the same time, it presents a more significant challenge to the network's capacity to discriminate between disparate features. discriminate between disparate features. We chose this type of sample (C, D sample) for a detailed local comparison, as  in Figure \ref{FIG:13}. In sample C, the difference features are present in both images corresponding to the blue box positions. Comparing the four networks, FC-Siam-Conc, IFNet, BIT, and Ours, for complex samples, Ours achieves the best results. Although there is still room for improvement, Ours performs better in terms of edge detail and number of targets compared to several other networks. The same problem appears in the sample D, where all three comparison networks show missing targets. Ours achieves optimal results not only for edge detail but also for disparity target recognition.

Samples E and F contain densely packed targets. In Sample E, the targets are not only dense but also small in size, and closely resemble the surrounding environment, which makes the edge information more blurred. From the processing results, our approach detects all targets, while networks like BIT and ChangeFormer exhibit different levels of target omission. The same issue also occurs in Sample F.

\begin{figure}
	\begin{center}
		\begin{tabular}{c}
			\includegraphics[height=6cm]{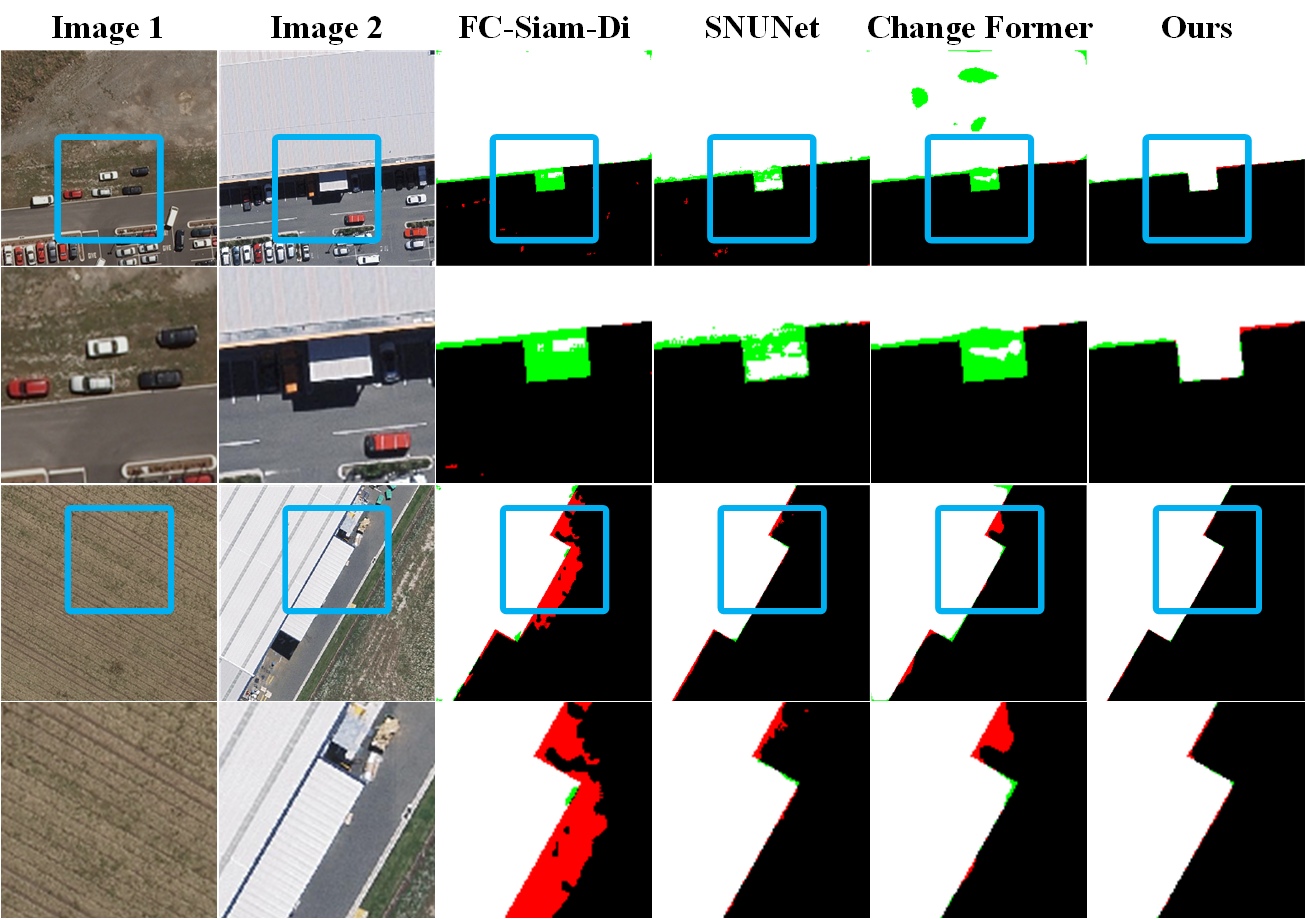}
		\end{tabular}
	\end{center}
	\caption
	{ \label{FIG:16}
		Detailed comparison chart of large targets.}
\end{figure}

\begin{figure}
	\begin{center}
		\begin{tabular}{c}
			\includegraphics[height=6cm]{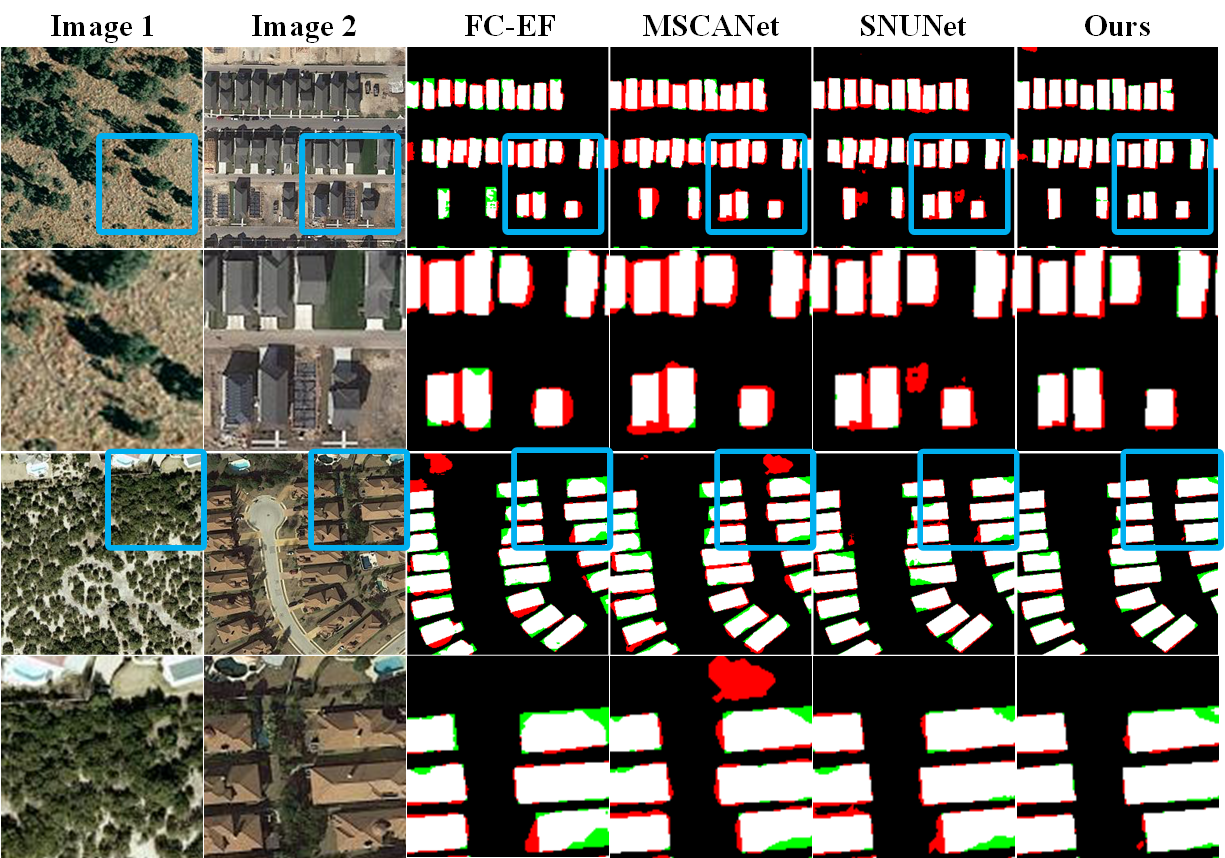}
		\end{tabular}
	\end{center}
	\caption
	{ \label{FIG:14}
		Detailed comparison chart of dense targets.}
\end{figure}

To highlight the change detection capability for dense targets, we have re-selected two sets of samples for local amplification. As in Figure \ref{FIG:14}, we select three networks, FC-EF, MSCANet, and SNUNet, for comparison with Ours. In the first sample, the dense arrangement of houses overlaps the shadows of the houses due to the influence of sunlight, resulting in blurred house boundaries. From the treatment results, Ours has the highest similarity to GT compared to the other three comparison methods, although Ours is also affected by environmental factors. During the processing of the second sample, all three comparison methods exhibit different levels of missing features and incorrect identifications due to tree occlusion, whereas Ours maximises the concentration of feature information and avoids incorrect identifications.

\begin{table*}[!t]
	\caption{Indicator results obtained for each method on the LEVIR-CD dataset. Red for best results, blue for second best results. All scores are described as percentages(\%).\label{tab1}}
	\centering
	\renewcommand\arraystretch{1.2}
	\scalebox{0.8}{
		\resizebox{\textwidth}{!}
		{
			\begin{tabular}{cccccc}
				\hline
				\textbf{Method} & \textbf{OA} & \textbf{F1} & \textbf{IoU} & \textbf{Precision} & \textbf{Recall} \\
				\hline
				FC-EF & 98.379 & 84.02 & 72.447 & 84.387 & 83.660 \\
				FC-Siam-Di & 98.714 & 87.169 & 77.256 & 88.626 & 85.759 \\
				FC-Siam-Conc & 98.688 & 87.254 & 77.389 & 86.374 & 88.151 \\
				DTCDSCN & 99.002 & 90.090 & 81.967 & 91.183 & 89.023 \\
				MSCANet & 98.859 & 88.665 & 79.638 & 89.786 & 87.571 \\
				IFNet & 98.949 & 89.319 & 80.699 & {\color[HTML]{0070C0} \textbf{92.568}} & 86.290 \\
				SNUNet & 98.665 & 86.317 & 75.928 & 90.325 & 82.650 \\
				BIT & 99.029 & 90.342 & 82.385 & 91.525 & 89.189 \\
				Change Former & 99.039 & 90.400 & 82.482 & 92.054 & 88.805 \\
				ICIF-Net & {\color[HTML]{0070C0} \textbf{99.122}} & {\color[HTML]{0070C0} \textbf{91.176}} & {\color[HTML]{0070C0} \textbf{83.851}} & 91.133 & {\color[HTML]{FF0000} \textbf{90.566}} \\
				DMINet & 99.07 & 90.71 & 82.99 & 92.52 & 89.95 \\
				AMTNet & 99.07 & 90.76 & 83.08 & 91.82 & 89.71 \\
				MFDS-Net & {\color[HTML]{FF0000} \textbf{99.154}} & {\color[HTML]{FF0000} \textbf{91.589}} & {\color[HTML]{FF0000} \textbf{84.483}} & {\color[HTML]{FF0000} \textbf{92.758}} & {\color[HTML]{0070C0} \textbf{90.449}}
				\\
				\hline
			\end{tabular}
	}}
\end{table*}

Table \ref{tab1} shows the results of the metrics obtained for each method on LEVIR-CD. From the results it appears that our network achieves optimal results on four  metrics. On the \textit{F1} metric, MFDS-Net scored 0.413 points higher than ICIF-Net. In the \textit{IoU} metric, MFDS-Net scored 0.632 points higher than ICIF-Net. The combined results show that MFDS-Net shows excellent processing capabilities on the LEVIR-CD dataset. Despite this, the MFDS-Net does not show a large performance advantage on some samples, which also indicates that there is potential for enhancing the performance of the MFDS-Net.

\subsubsection{WHU-CD}

\begin{figure*}
	\begin{center}
		\begin{tabular}{c}
			\includegraphics[height=6.8cm]{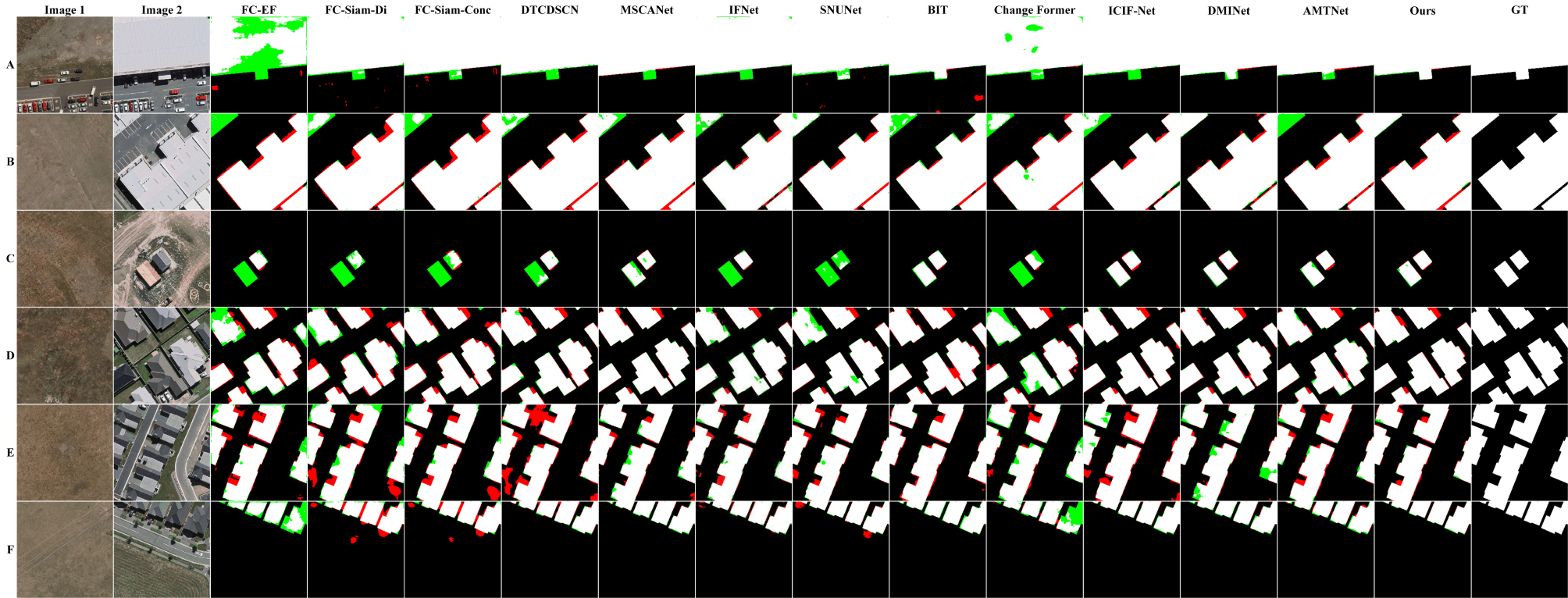}
		\end{tabular}
	\end{center}
	\caption
	{ \label{FIG:15}
		Graph of prediction results obtained by each method on the WHU-CD dataset.}
\end{figure*}

Figure \ref{FIG:15} shows the experimental results of each method on the WHU-CD dataset. In the processing of sample A Ours shows excellent ability in feature continuity and edge feature processing. In contrast, ICIF-Net, AMTNet and others show different degrees of recognition problems. In sample B, a large number of shadows appear around the buildings due to the exposure to sunlight. The FC-Siam-Conc and BIT methods handle the shadows poorly and show obvious detection deficiency of the building in the upper left corner of the sample. In contrast, Ours, although has certain limitations in processing, does not show missing phenomena in building recognition.

Sample A is selected in Figure \ref{FIG:16} and an additional set of samples is included for local enlargement comparison, to emphasize the processing capability for large buildings. In the first sample, due to the shadow obstruction of the protruding parts of the building, the FC-Siam-Di, SNUNet, and Change Former networks have obvious deficiencies in building detection, while Ours effectively avoids the influence of shadows and achieves satisfactory results in edge processing. In the second sample, due to the presence of interference at the corner position, the three comparison networks are significantly affected, while Ours can effectively distinguish and output the range of detection targets, and achieves results closest to the GT in edge processing.

\begin{figure}
	\begin{center}
		\begin{tabular}{c}
			\includegraphics[height=6cm]{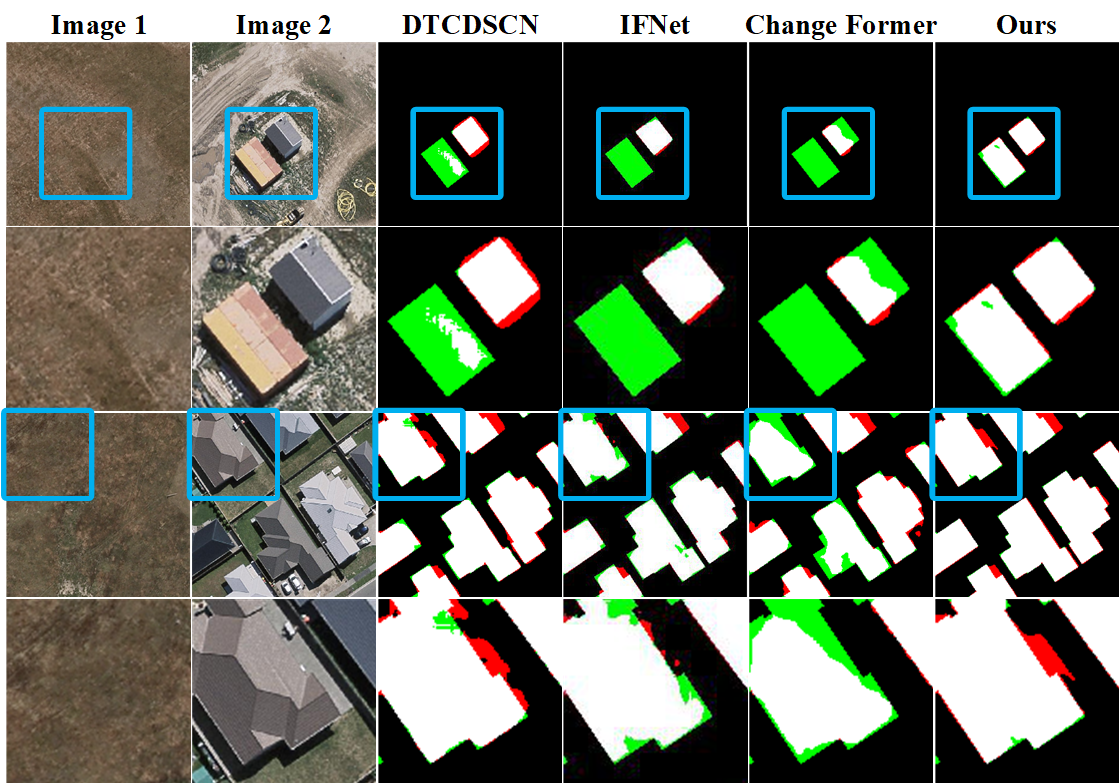}
		\end{tabular}
	\end{center}
	\caption
	{ \label{FIG:17}
		Diverse target details enlarged for comparison.}
\end{figure}

To emphasize the capability to detect various targets, we select Sample C and Sample D for a local zoom-in comparison, as in Figure \ref{FIG:17}. The two targets in sample C have two completely different colours, and the three networks DTCDSCN, IFNet, and Change Former are clearly missing the recognition and feature learning of the buildings. Ours, on the other hand, accomplished the detection of the two different coloured buildings and showed the best results in edge processing. In sample D, the houses are in different colours. Although Ours also showed some errors, the targets we detected were not missing compared to the other three methods.

The objects in the sample F have complex structures and colors. FC-EF, Change Former's segmentation of the house on the right is obviously missing features, and the edge judgment is obviously insufficient. However, SNUNet, etc. are affected by light, and output shadows as detection targets. Ours achieved nearly the same results as GT.

\begin{figure}
	\begin{center}
		\begin{tabular}{c}
			\includegraphics[height=6cm]{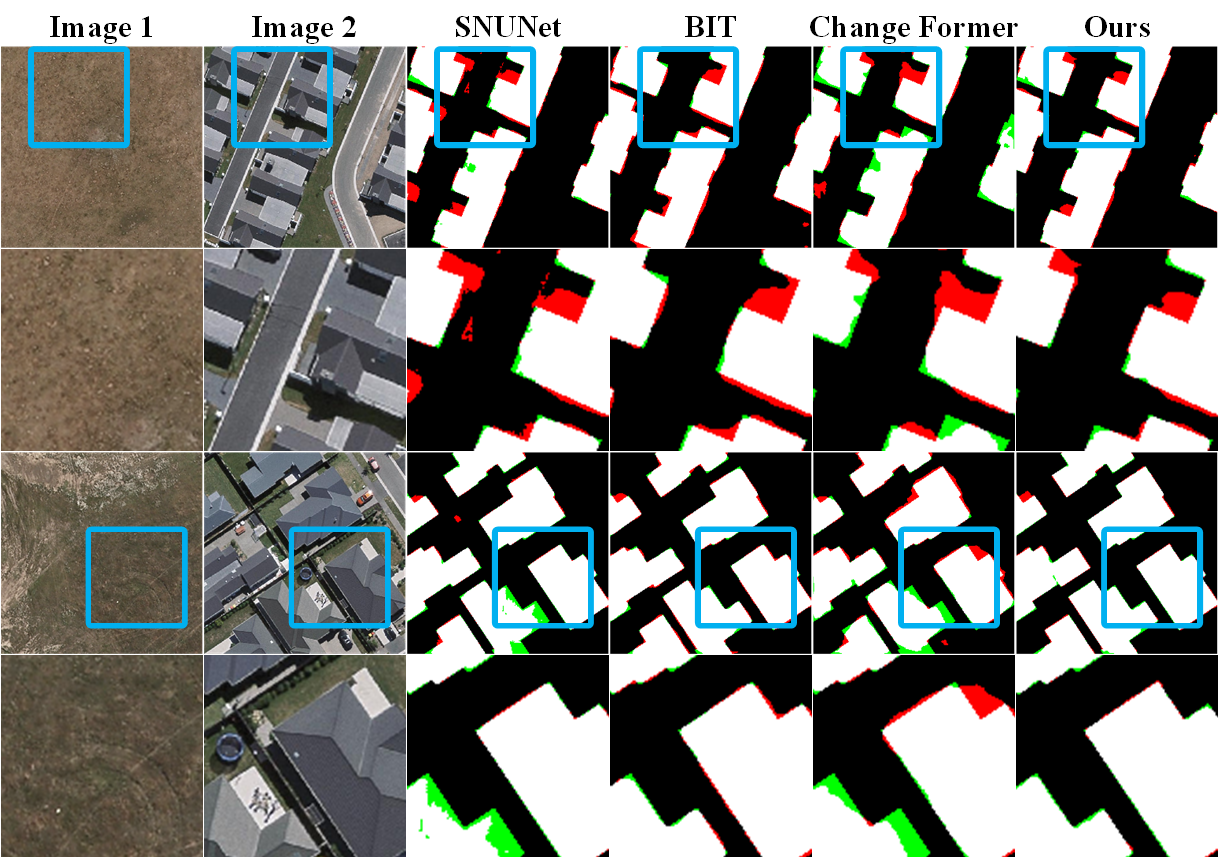}
		\end{tabular}
	\end{center}
	\caption
	{ \label{FIG:18}
		Detail comparison chart of complex samples.}
\end{figure}

In Figure \ref{FIG:18}, the buildings in both sets of samples have relatively complex structures, especially in the first set of samples where the houses and surrounding roads have the same color and texture structure. From the processing of the first set of samples, including Ours, the four methods all have some detection errors due to environmental factors such as lighting and shadows, but Ours has the least impact according to the comparison of magnified details. From the magnification results obtained by various comparison methods in the second set of samples, Ours achieves relatively good results, but Ours is better in handling details.

\begin{table*}[!t]
	\caption{Indicator results obtained for each method on the WHU-CD dataset. Red for best results, blue for second best results. All scores are described as percentages(\%).\label{tab2}}
	\centering
	\renewcommand\arraystretch{1.2}
	\scalebox{0.8}{
		\resizebox{\textwidth}{!}
		{
			\begin{tabular}{cccccc}
				\hline
				\textbf{Method} & \textbf{OA} & \textbf{F1} & \textbf{IoU} & \textbf{Precision} & \textbf{Recall} \\
				\hline
				FC-EF & 98.015 & 77.642 & 63.452 & 84.608 & 71.742 \\
				FC-Siam-Di & 98.060 & 79.973 & 66.627 & 79.783 & 80.165 \\
				FC-Siam-Conc & 98.454 & 84.125 & 72.596 & 83.070 & 85.194 \\
				DTCDSCN & 99.054 & 90.031 & 81.870 & 91.459 & 88.548 \\
				MSCANet & 98.954 & 88.619 & 79.564 & 93.164 & 84.497 \\
				IFNet & 98.832 & 83.405 & 71.525 & {\color[HTML]{FF0000} \textbf{96.914}} & 73.196 \\
				SNUNet & 98.912 & 88.339 & 79.114 & 91.340 & 85.530 \\
				BIT & 98.809 & 87.471 & 77.732 & 88.707 & 86.269 \\
				Change Former & 98.758 & 86.882 & 76.806 & 88.495 & 85.326 \\
				ICIF-Net & 99.13 & 90.766 & 83.093 & 92.927 & 88.702 \\
				DMINet & {\color[HTML]{0070C0} \textbf{99.19}} & {\color[HTML]{0070C0} \textbf{91.49}} & {\color[HTML]{0070C0} \textbf{84.31}} & 92.65 & {\color[HTML]{0070C0} \textbf{90.35}} \\
				AMTNet & 99.1 & 90.57 & 82.62 & 91.11 & 89.97 \\
				MFDS-Net & {\color[HTML]{FF0000} \textbf{99.324}} & {\color[HTML]{FF0000} \textbf{92.938}} & {\color[HTML]{FF0000} \textbf{86.807}} & {\color[HTML]{0070C0} \textbf{93.606}} & {\color[HTML]{FF0000} \textbf{92.279}}
				\\
				\hline
			\end{tabular}
	}}
\end{table*}

As in Table \ref{tab2}, MFDS-Net achieved the best results in four metrics: \textit{OA}, \textit{F1}, \textit{IOU}, and \textit{Recall}, with 1.448 points higher than DMINet in \textit{F1}, and 2.497 points higher than DMINet in\textit{ IoU}. However, there is still a gap between MFDS-Net and IFNet in terms of Precision, which shows that MFDS-Net still has room for performance improvement.

The combined experimental results show that MFDS-Net can still show a high CD ability for samples with more detailed information. When faced with samples with diverse targets, MFDS-Net demonstrated a fitting ability higher than current mainstream methods, but there is still room for improvement in handling the effects of lighting.

\begin{figure*}
	\begin{center}
		\begin{tabular}{c}
			\includegraphics[height=6.8cm]{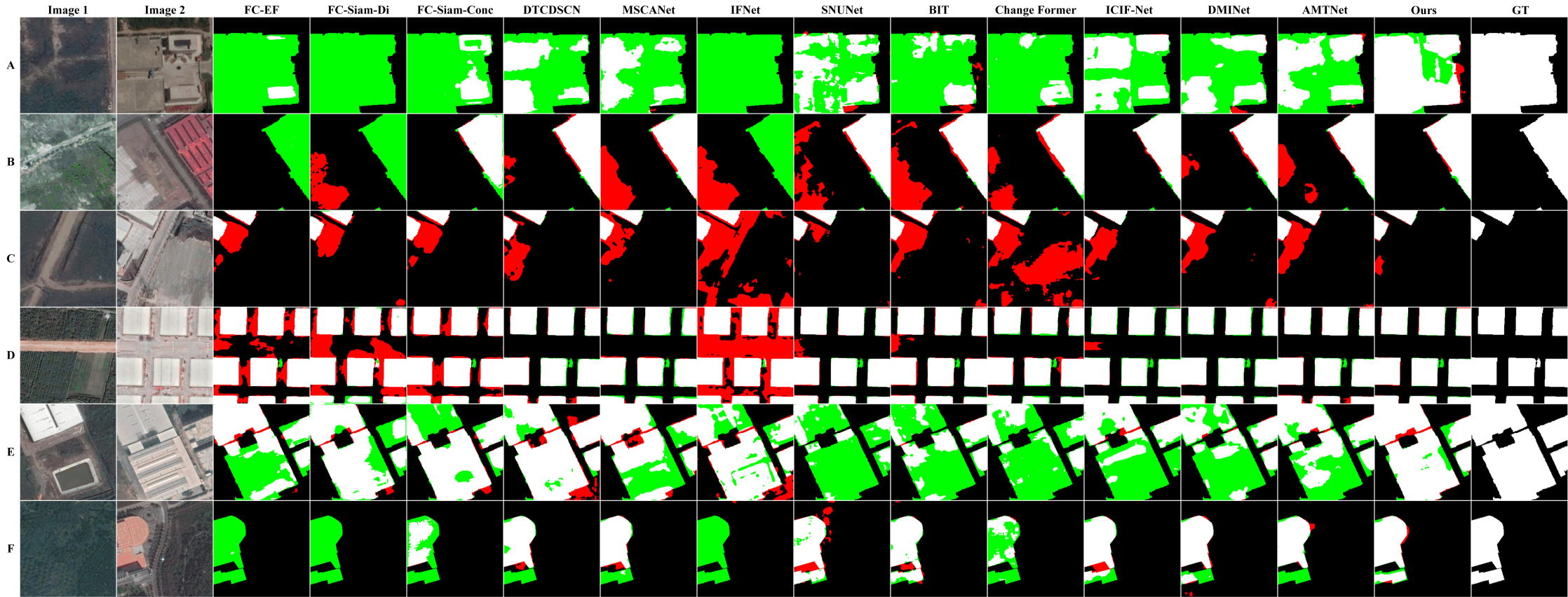}
		\end{tabular}
	\end{center}
	\caption
	{ \label{FIG:19}
		Graph of prediction results obtained by each method on the GZ-CD dataset.}
\end{figure*}
\subsubsection{GZ-CD}
\begin{figure}
	\begin{center}
		\begin{tabular}{c}
			\includegraphics[height=6cm]{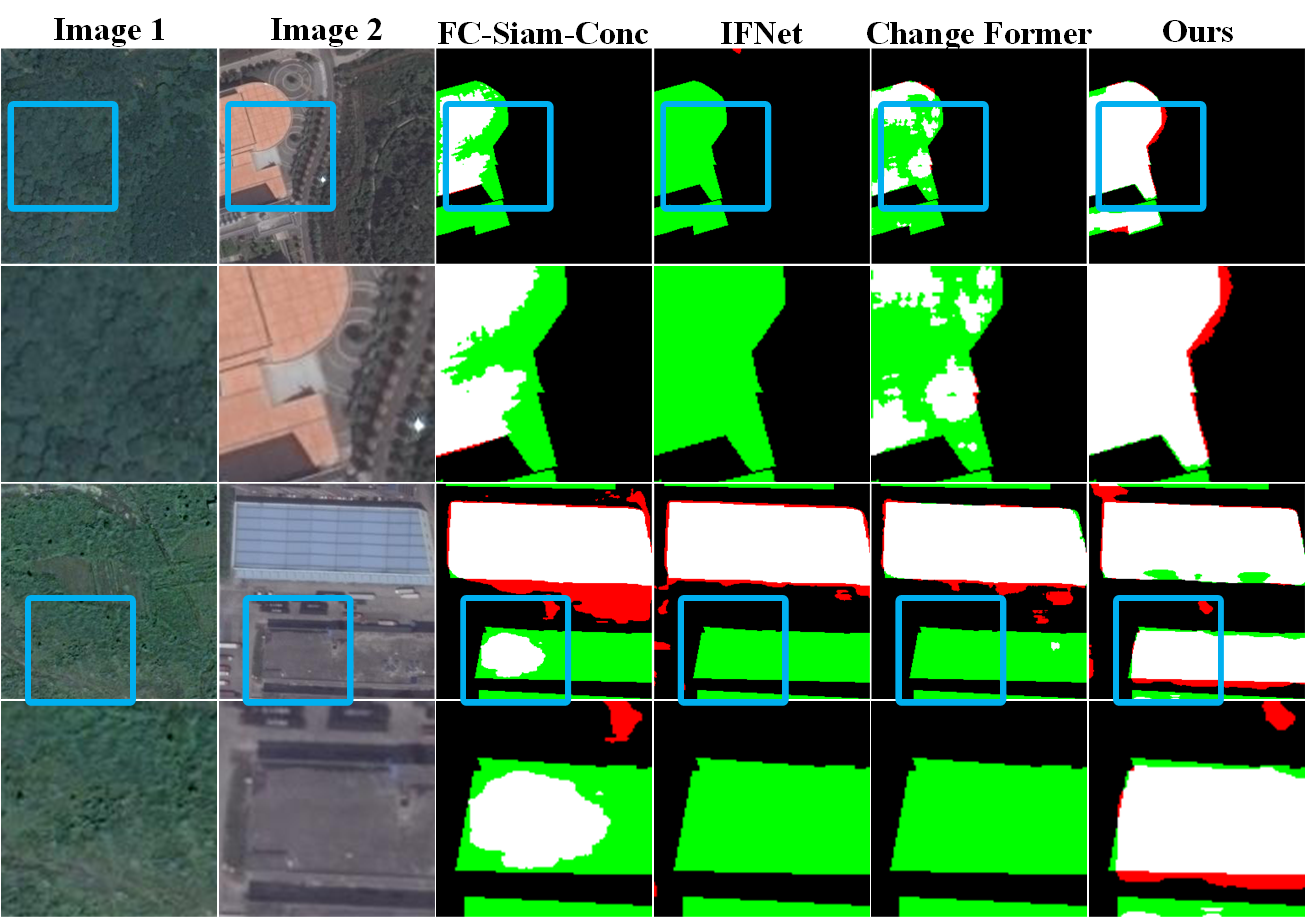}
		\end{tabular}
	\end{center}
	\caption
	{ \label{FIG:20}
		Diversified target details zoom in and compare.}
\end{figure}

The outcomes of each approach on the GZ-CD are illustrated in Figure \ref{FIG:19}. BIT, Change Former and other networks are clearly deficient in determining the target edges in samples A and C. Although Ours exhibits shortcomings in maintaining feature continuity when handling sample A, it still outperforms the other methods. In contrast, Ours achieved the closest results to GT in the sample C, with the best performance in edge processing as well as in the treatment of distractors.

Compared to samples A and C, the detection targets in sample B have different colors, and the EC-EF, FC-Siam-Di, and IFNet networks incorrectly classify the target as environmental features. Especially for IFNet, not only is the detection target misjudged, but the environment in the lower left corner is also misclassified as a detection target. The same issue also occurs in networks such as BIT, Change Former and AMTNet.

In sample F, the color of the differential target is different from that of general buildings, which makes it difficult for networks such as FC-Siam-Di, IFNet, BIT to recognize it, which also indirectly indicates that these methods have insufficient performance in feature fitting.

To highlight the treatment of diverse variance features, we selected the sample F and supplemented them with a new set of samples for local enlargement, as in Figure \ref{FIG:20}. In another sample, the two targets have different colours and the three contrast networks are only able to achieve processing for the blue target. In contrast, Ours is able to separate the targets from the environment more completely, although the processing of the target edges below Image2 is less refined. This also demonstrates the high feature-fitting capability of our network.

\begin{table*}[!t]
	\caption{Indicator results obtained for each method on the GZ-CD dataset. Red for best results, blue for second best results. All scores are described as percentages(\%).\label{tab3}}
	\centering
	\renewcommand\arraystretch{1.2}
	\scalebox{0.8}{
		\resizebox{\textwidth}{!}
		{
			\begin{tabular}{cccccc}
				\hline
				\textbf{Method} & \textbf{OA} & \textbf{F1} & \textbf{IoU} & \textbf{Precision} & \textbf{Recall} \\
				\hline
				FC-EF & 94.888 & 71.129 & 55.194 & 77.488 & 65.735 \\
				FC-Siam-Di & 94.014 & 65.456 & 48.650 & 73.177 & 59.208 \\
				FC-Siam-Conc & 95.682 & 76.244 & 61.608 & 80.588 & 72.344 \\
				DTCDSCN & 96.833 & 82.719 & 70.531 & 86.649 & 79.131 \\
				MSCANet & 96.398 & 80.663 & 67.593 & 83.037 & 78.421 \\
				IFNet & 96.917 & 82.152 & 69.711 & {\color[HTML]{FF0000} \textbf{92.194}} & 74.083 \\
				SNUNet & 97.069 & 84.250 & 72.786 & 86.824 & {\color[HTML]{0070C0} \textbf{81.824}} \\
				BIT & 96.310 & 80.233 & 66.992 & 82.401 & 78.177 \\
				Change Former & 95.531 & 73.657 & 58.300 & 84.591 & 65.227 \\
				ICIF-Net & {\color[HTML]{0070C0} \textbf{97.289}} & {\color[HTML]{0070C0} \textbf{85.09}} & {\color[HTML]{0070C0} \textbf{74.049}} & {\color[HTML]{0070C0} \textbf{89.904}} & 80.764 \\
				DMINet & 96.77 & 81.98 & 69.46 & 87.92 & 76.79 \\
				AMTNet & 96.831 & 82.38 & 70.03 & 87.98 & 77.44 \\
				MFDS-Net & {\color[HTML]{FF0000} \textbf{97.581}} & {\color[HTML]{FF0000} \textbf{86.377}} & {\color[HTML]{FF0000} \textbf{76.021}} & 88.918 & {\color[HTML]{FF0000} \textbf{83.978}}
				\\
				\hline
			\end{tabular}
	}}
\end{table*}

Table \ref{tab3} shows that MFDS-Net achieves the highest scores in four metrics for all the methods involved in the experiment. MFDS-Net shows a higher performance on the GZ-CD than the current mainstream methods. However, MFDS-Net still has potential for improvement, for example, in the processing of sample A in Figure \ref{FIG:19}, MFDS-Net underperforms in maintaining continuity of large target features.

\begin{figure}
	\begin{center}
		\begin{tabular}{c}
			\includegraphics[height=4cm]{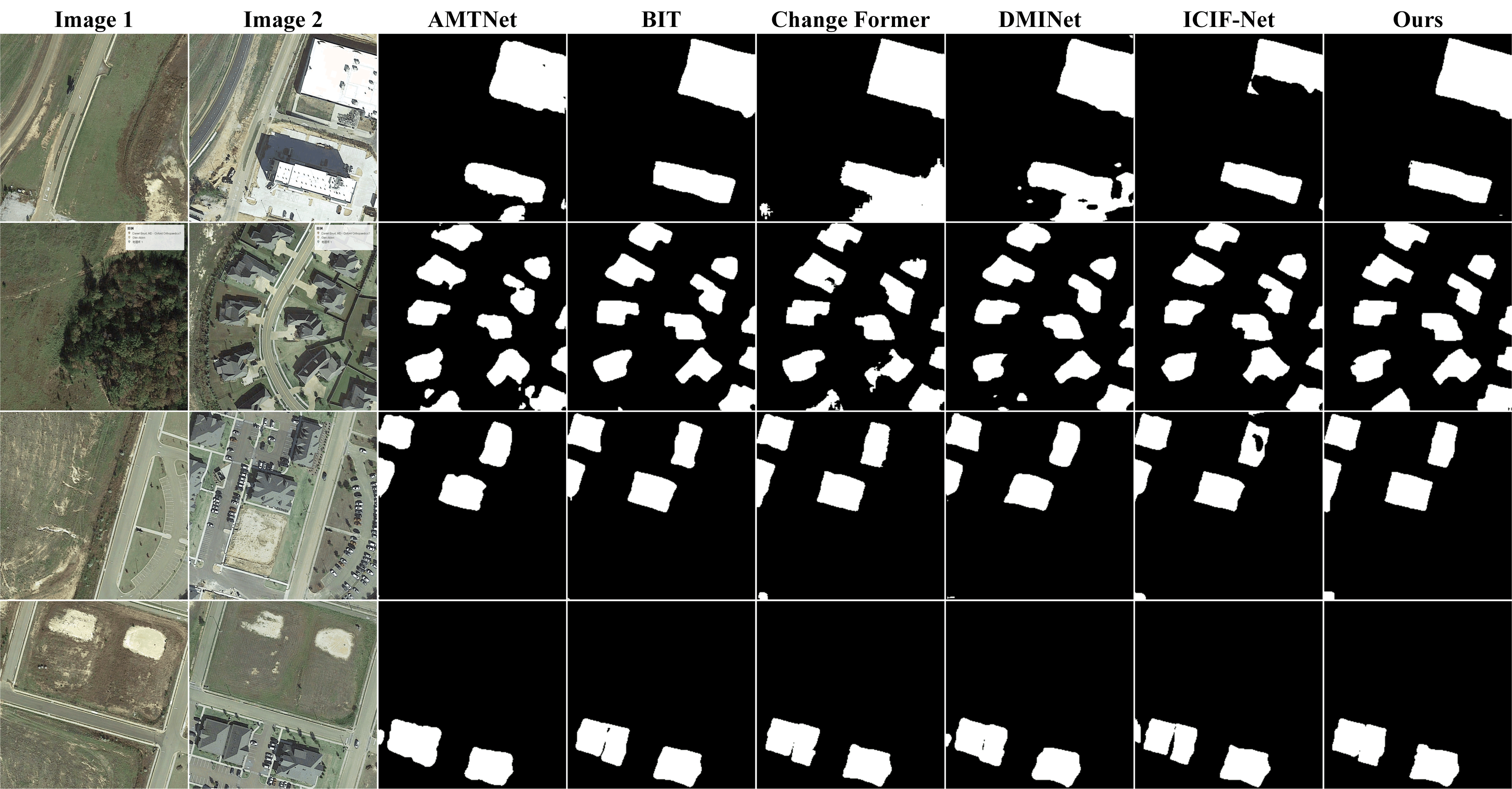}
		\end{tabular}
	\end{center}
	\caption
	{ \label{njwh} The results of models trained on the LEVIR dataset, including AMTNet, BIT, ChangeFormer, ICIF-Net, DMINet, and Ours, in practical applications.}
\end{figure}
\subsubsection{Parametric quantities and floating point calculations}

\begin{table*}[!t]
	\caption{Parametric quantities and floating point calculations.\label{tab4}}
	\centering
	\renewcommand\arraystretch{1.2}
	\resizebox{\textwidth}{!}
	{
		\begin{tabular}{c|cc|cc|cc|cc}
			\hline
			\multirow{2}{*}{\textbf{Method}} & \multicolumn{2}{c}{\textbf{Complexity}} & \multicolumn{2}{c}{\textbf{LEVIR-CD}} & \multicolumn{2}{c}{\textbf{WHU-CD}} & \multicolumn{2}{c}{\textbf{GZ-CD}} \\ \cline{2-9} 
			& Params(M) & FLOPs(G) & F1 & IoU & F1 & IoU & F1 & IoU \\
			\hline
			FC-EF & 1.35 & 3.57 & 84.022 & 72.447 & 77.642 & 63.452 & 71.129 & 55.194 \\
			FC-Siam-Di & 1.35 & 4.72 & 87.169 & 77.256 & 79.973 & 66.627 & 65.456 & 48.650 \\
			FC-Siam-Conc & 1.55 & 5.32 & 87.254 & 77.389 & 84.125 & 72.596 & 76.244 & 61.608 \\
			DTCDSCN & 41.07 & 13.21 & 90.090 & 81.967 & 90.031 & 81.870 & 82.719 & 70.531 \\
			MSCANet & 16.59 & 14.70 & 88.665 & 79.638 & 88.619 & 79.564 & 80.663 & 67.593 \\
			IFNet & 50.71 & 82.35 & 89.319 & 80.699 & 83.405 & 71.525 & 82.152 & 69.711 \\
			SNUNet & 12.03 & 54.88 & 86.317 & 75.928 & 88.339 & 79.114 & 84.250 & 72.786 \\
			BIT & 3.55 & 10.59 & 90.342 & 82.385 & 87.471 & 77.732 & 80.233 & 66.992 \\
			Change Former & 41.03 & 202.87 & 90.400 & 82.482 & 86.882 & 76.806 & 73.657 & 58.300 \\
			ICIF-Net & 25.83 & 25.27 & 91.176 & 83.851 & 90.766 & 83.093 & 85.09 & 74.049 \\
			DMINet & 6.24 & 14.55 & 90.71 & 82.99 & 91.49 & 84.31 & 81.98 & 69.46 \\
			AMTNet & 24.67 & 86.23 & 90.76 & 83.08 & 90.57 & 82.62 & 82.38 & 70.03 \\
			MFDS-Net & 25.67 & 1.76 & 91.589 & 84.483 & 92.938 & 86.807 & 86.377 & 76.021
			\\
			\hline
		\end{tabular}
	}%
\end{table*}

Table \ref{tab4} shows the number of parameters for each method. The parameters of MFDS-Net are lower than those of IFNet, DTCDSCN, ICIF-Net, and Change Former. Although MFDS-Net is higher than FC-EF, BIT and other light networks in terms of parameter quantity, it also obtains better results. MFDS-Net has reached the lowest in floating-point data calculation, which is lower than all current mainstream methods.

Additionally, to validate the predictive accuracy of our model on changing targets in real scenarios, we utilized Google Earth to obtain images of the Oxford area in Mississippi, USA, from 2012 and 2019. In Figure \ref{njwh}, we compared MFDS-Net with BIT, Change Former, ICIF-Net, DMINet, and AMTNet.

\subsection{Ablation Experiment}\label{subsec44}

The details of the ablation experiment are described in this section. As in Table \ref{tab5}, the MDPM used in the network is removed and the ablation model is named MFDS-Net-1. The depth supervision mechanism is removed in MFDS-Net-2. To verify the effectiveness of the GSEM module, it is removed from MFDS-Net-3. Subsequently, we eliminate the DFIM in the original network and fuse the three features directly using the form of channel splicing and processing with convolutional blocks, which is named MFDS-Net-4. Finally, in MFDS-Net-5, DO-Conv is replaced with conventional convolution. We validate the module performance on three datasets and compare it with the original network.

\begin{table*}[!t]
	\caption{Ablation Model Form.\label{tab5}}
	\centering
	\renewcommand\arraystretch{1.2}
	\scalebox{0.8}{
		\resizebox{\textwidth}{!}
		{
			\begin{tabular}{cccccc}
				\hline
				\textbf{Method} & \textbf{MDPM} & \textbf{Deep supervision} & \multicolumn{1}{l}{\textbf{GSEM}} & \textbf{DFIM} & \textbf{DO-Conv} \\ \hline
				MFDS-Net-1 & \usym{1F5F4} & \usym{1F5F8} & \usym{1F5F8} & \usym{1F5F8} & \usym{1F5F8} \\
				MFDS-Net-2 & \usym{1F5F8} & \usym{1F5F4} & \usym{1F5F8} & \usym{1F5F8} & \usym{1F5F8} \\
				MFDS-Net-3 & \usym{1F5F8} & \usym{1F5F8} & \usym{1F5F4} & \usym{1F5F8} & \usym{1F5F8} \\
				MFDS-Net-4 & \usym{1F5F8} & \usym{1F5F8} & \usym{1F5F8} & \usym{1F5F4} & \usym{1F5F8} \\
				MFDS-Net-5 & \usym{1F5F8} & \usym{1F5F8} & \usym{1F5F8} & \usym{1F5F8} & \usym{1F5F4} \\
				MFDS-Net & \usym{1F5F8} & \usym{1F5F8} & \usym{1F5F8} & \usym{1F5F8} & \usym{1F5F8} \\ \hline
			\end{tabular}
	}}
\end{table*}

\subsubsection{Parameter adjustment}

As in Table \ref{tab6}, we compare the performance of the network with different learning rates. The results show that the training performance and convergence of the network with a learning rate of 0.0001 are significantly inferior to those with a learning rate of 0.001. With 200 rounds as the upper limit, on the LEVIR-CD dataset, the network with a learning rate of 0.001 reached its highest score at 132 rounds, with an F1 score of 91.589; while with a learning rate of 0.0001, even though it reached 189 rounds, the F1 score was only 91.305.

\begin{table*}[!t]
	\caption{Learning Rate Adjustment Form.\label{tab6}}
	\centering
	\renewcommand\arraystretch{2}
	\scalebox{1}{
		\resizebox{\textwidth}{!}
		{
			\begin{tabular}{c|ccc|ccc|lll}
				\hline
				\multirow{2}{*}{\textbf{Learning Rate}} & \multicolumn{3}{c|}{\textbf{LEVIR-CD (F1\%)}} & \multicolumn{3}{c|}{\textbf{WHU-CD (F1\%)}} & \multicolumn{3}{c}{\textbf{GZ-CD (F1\%)}} \\ \cline{2-10} 
				& \multicolumn{1}{c|}{\textbf{100epoch}} & \multicolumn{1}{c|}{\textbf{150epoch}} & \textbf{200epoch} & \multicolumn{1}{c|}{\textbf{100epoch}} & \multicolumn{1}{c|}{\textbf{150epoch}} & \textbf{200epoch} & \multicolumn{1}{c|}{\textbf{100epoch}} & \multicolumn{1}{c|}{\textbf{150epoch}} & \multicolumn{1}{c}{\textbf{200epoch}} \\ \hline
				\textbf{0.001} & \multicolumn{1}{c|}{91.454(100)} & \multicolumn{1}{c|}{91.589(132)} & 91.589(132) & \multicolumn{1}{c|}{91.106(99)} & \multicolumn{1}{c|}{91.812(139)} & 92.938(189) & \multicolumn{1}{l|}{84.984(95)} & \multicolumn{1}{l|}{86.023(143)} & 86.377(168) \\
				\textbf{0.0001} & \multicolumn{1}{l|}{90.389(71)} & \multicolumn{1}{l|}{90.877(120)} & \multicolumn{1}{l|}{91.305(189)} & \multicolumn{1}{l|}{84.705(76)} & \multicolumn{1}{c|}{88.708(119)} & \multicolumn{1}{l|}{90.651(197)} & \multicolumn{1}{l|}{81.922(91)} & \multicolumn{1}{l|}{85.171(132)} & 86.581(184)\\
				\hline
			\end{tabular}
	}}%
\end{table*}

\subsubsection{LEVIR-CD}

\begin{figure}
	\begin{center}
		\begin{tabular}{c}
			\includegraphics[height=3.5cm]{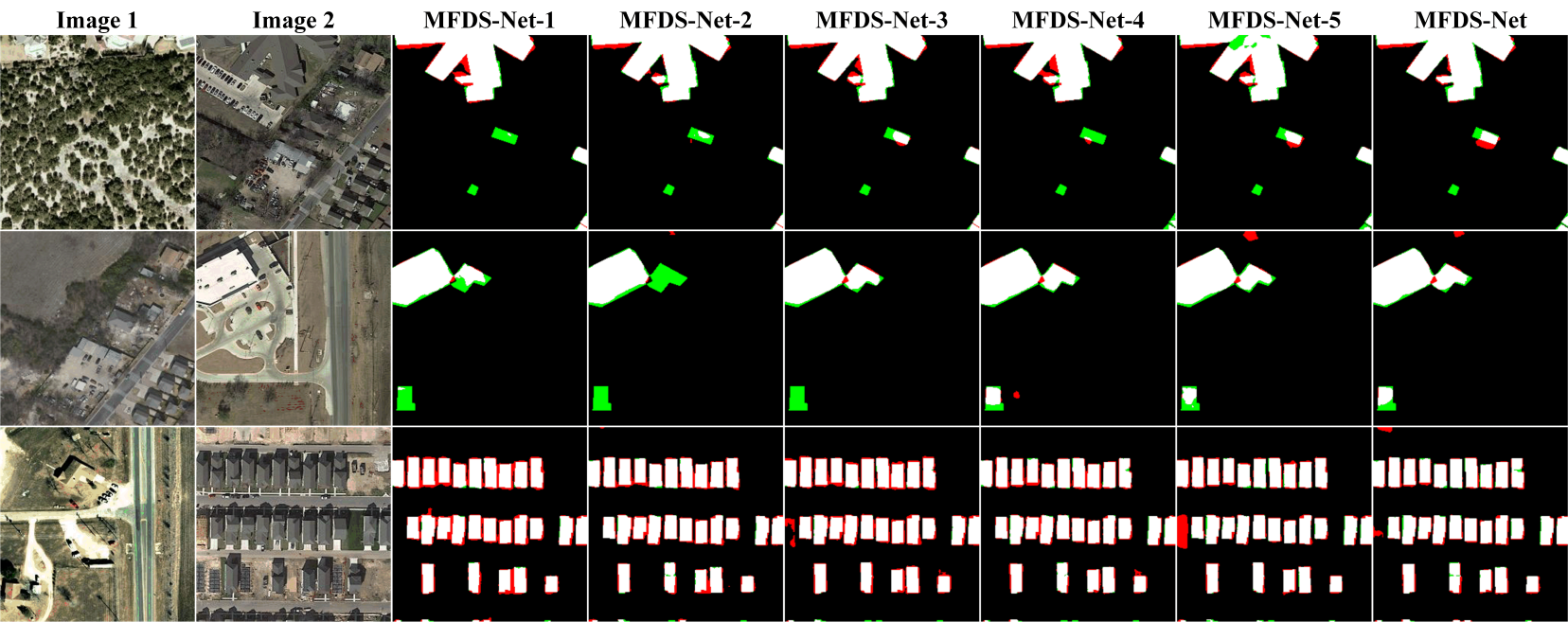}
		\end{tabular}
	\end{center}
	\caption
	{ \label{FIG:21}
		The results of the ablation experiment on the LEVIR-CD dataset.}
\end{figure}

Figure \ref{FIG:21} shows the experimental results on LEVIR-CD. The overall results show that MFDS-Net-1 and MFDS-Net-2 exhibit a large gap between the results on edge feature processing and the original network. This illustrates the high contribution of MFEA and deep supervision mechanisms to the ability to focus on weak features.

\begin{table}[!t]
	\caption{Indicators obtained from ablation experiments on the LEVIR-CD dataset\label{tab7}}
	\centering
	\renewcommand\arraystretch{1.2}
	\scalebox{0.5}{
		\resizebox{\textwidth}{!}
		{
			\begin{tabular}{cccccc}
				\hline
				\textbf{Method} & \textbf{OA} & \textbf{F1} & \textbf{IoU} & \textbf{Precision} & \textbf{Recall} \\
				\hline
				MFDS-Net-1 & 99.108 & 91.240 & 83.891 & 91.333 & 91.148 \\
				MFDS-Net-2 & 99.065 & 90.745 & 83.059 & 91.562 & 89.944 \\
				MFDS-Net-3 & 99.120 & 91.334 & 84.050 & 91.619 & 91.050 \\
				MFDS-Net-4 & 99.060 & 90.669 & 82.931 & 91.747 & 89.616 \\
				MFDS-Net-5 & 99.128 & 91.270 & 83.942 & 93.090 & 89.520 \\
				MFDS-Net & 99.154 & 91.589 & 84.483 & 92.758 & 90.449
				\\
				\hline
			\end{tabular}
	}}
\end{table}

Table \ref{tab7} shows the metric results for the ablation models and the original network on LEVIR-CD. Although the original network was not optimal in terms of individual metrics, the overall results show that all five ablation models achieved lower results than the original network.

\subsubsection{WHU-CD}

\begin{figure}
	\begin{center}
		\begin{tabular}{c}
			\includegraphics[height=3.5cm]{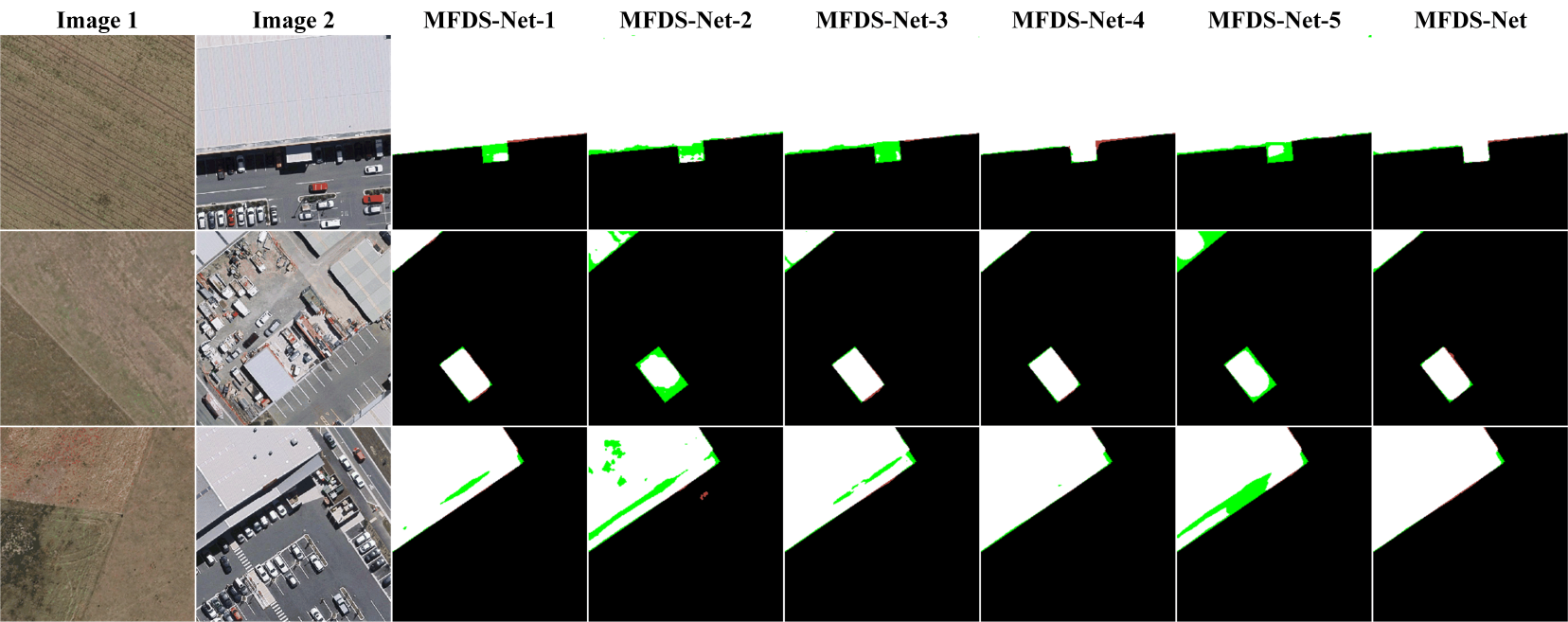}
		\end{tabular}
	\end{center}
	\caption
	{ \label{FIG:22}
		The results of the ablation experiment on the WHU-CD dataset.}
\end{figure}

Figure \ref{FIG:22} displays the outcomes achieved by the ablation models in WHU-CD. Except for MFDS-Net-4, which is similar to the results obtained by the original model, all the other four ablation models have certain deficiencies in terms of feature continuity. From the perspective of detailed features, MFDS-Net-4 still does not handle edge features as well as the original network. This also proves from the side that the modules corresponding to the five ablation models contribute to the performance of the network.

\begin{table}[!t]
	\caption{Indicators obtained from ablation experiments on the WHU-CD dataset\label{tab8}}
	\centering
	\renewcommand\arraystretch{1.2}
	\scalebox{0.5}{
		\resizebox{\textwidth}{!}
		{
			\begin{tabular}{cccccc}
				\hline
				\textbf{Method} & \textbf{OA} & \textbf{F1} & \textbf{IoU} & \textbf{Precision} & \textbf{Recall} \\
				\hline
				MFDS-Net-1 & 99.149 & 92.066 & 85.298 & 92.401 & 91.733 \\
				MFDS-Net-2 & 98.848 & 89.832 & 81.541 & 89.688 & 89.976 \\
				MFDS-Net-3 & 99.181 & 91.901 & 84.094 & 92.932 & 90.893 \\
				MFDS-Net-4 & 99.112 & 91.639 & 84.569 & 92.336 & 90.953 \\
				MFDS-Net-5 & 99.092 & 91.477 & 84.291 & 92.701 & 90.248 \\
				MFDS-Net   & 99.324 & 92.938 & 86.807 & 93.606 & 92.279
				\\
				\hline
			\end{tabular}
	}}
\end{table}

The metrics achieved by the ablation model are presented in Table \ref{tab8}. Combined with the subjective analysis, the deep supervision mechanism plays an important role in for the training of WHU-CD. GSEM enhances the global relevance of high-level semantic information from a global perspective and improves the overall performance of MFDS-Net. Of course, other modules also make important contributions to the network.

\subsubsection{GZ-CD}

\begin{figure}
	\begin{center}
		\begin{tabular}{c}
			\includegraphics[height=3.5cm]{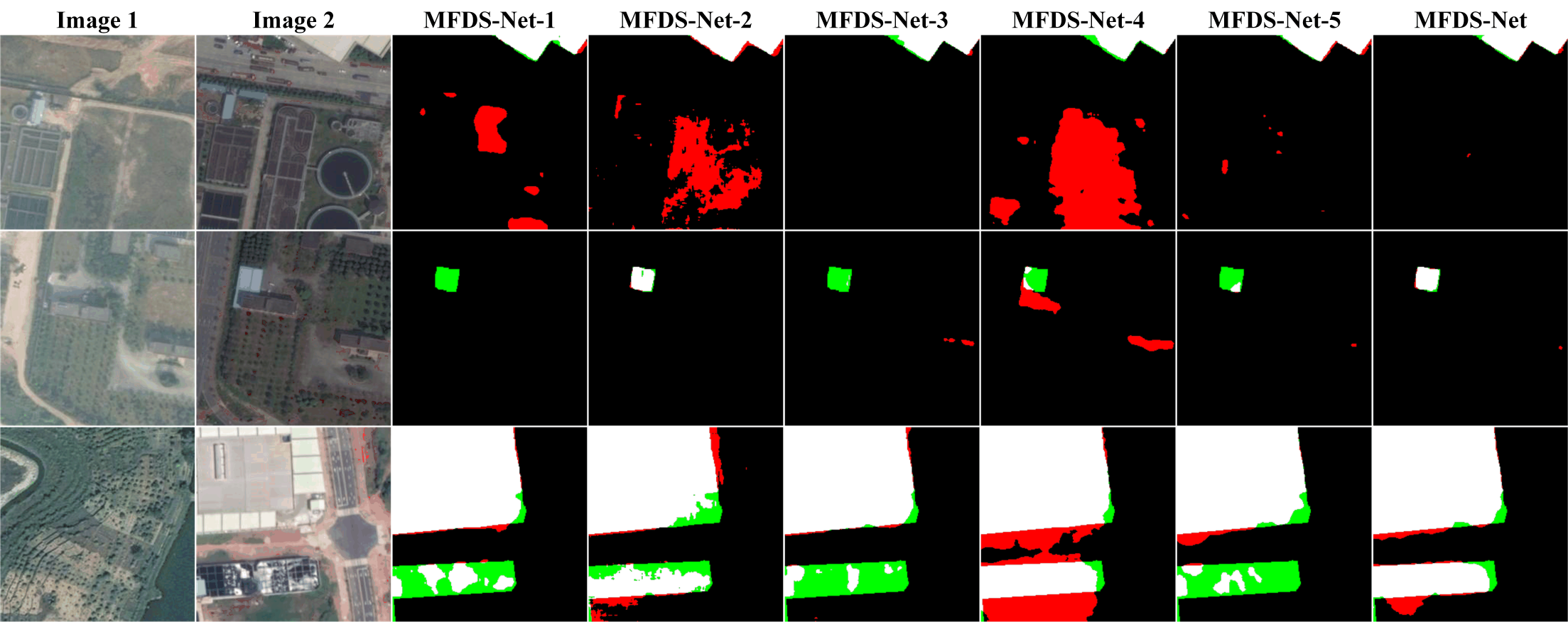}
		\end{tabular}
	\end{center}
	\caption
	{ \label{FIG:23}
		The results of the ablation experiment on the GZ-CD dataset.}
\end{figure}

Combining Figure \ref{FIG:23} with Table \ref{tab9}, the results obtained by the ablation model on the GZ-CD differ significantly from those of the original network, thus indirectly illustrating the importance of the module used by MFDS-Net. The detection of the second set of samples by MFDS-Net-5 shows missing features, a problem that is also present in the third set of samples. This also indicates that DO-Conv has some improvement in the training effect of the network.

\begin{table}[!t]
	\caption{Indicators obtained from ablation experiments on the GZ-CD dataset\label{tab9}}
	\centering
	\renewcommand\arraystretch{1.2}
	\scalebox{0.5}{
		\resizebox{\textwidth}{!}
		{
			\begin{tabular}{cccccc}
				\hline
				\textbf{Method} & \textbf{OA} & \textbf{F1} & \textbf{IoU} & \textbf{Precision} & \textbf{Recall} \\
				\hline
				MFDS-Net-1 & 97.496 & 86.039 & 75.498 & 87.086 & 85.016 \\
				MFDS-Net-2 & 96.916 & 84.036 & 72.467 & 83.332 & 84.752 \\
				MFDS-Net-3 & 97.429 & 86.084 & 75.567 & 87.212 & 84.984 \\
				MFDS-Net-4 & 97.638 & 86.07 & 75.546 & 89.266 & 83.095 \\
				MFDS-Net-5 & 97.377 & 85.997 & 75.434 & 88.01 & 84.073 \\
				MFDS-Net & 97.581 & 86.377 & 76.021 & 88.918 & 83.978
				\\
				\hline
			\end{tabular}
	}}
\end{table}

\section{Limitations and future work }\label{sec5}

MFDS-Net has shown results above the mainstream methods in terms of feature continuity, detection of target edge processing, and complex environment processing. Nevertheless, MFDS-Net is still not optimal in terms of performance. The results shown in some of the samples are not clearly advantageous compared to the mainstream methods. In addition, the results are not optimal in some of the indicators, which indicates that the MFDS-Net still needs to be improved in terms of stability and robustness. In the future, we will consider further compressing the network model and reducing redundancy, so as to further improve the performance of the network.

In addition, there is more room to explore the general applicability of MFDS-Net, and we will conduct experiments on more kinds of CD datasets in the future. In the future, we will conduct experiments on more kinds of CD datasets. In order to explore the cross-domain applicability of the network, we will also explore the field of remote sensing segmentation, medical image segmentation, semantic segmentation and other fields in the future.

\section{Summary}\label{sec6}

This paper proposes the MFDS-Net, a new network for coping with remote sensing CD tasks. MFDS-Net mainly responds to the current problems faced in CD tasks such as difficulties in processing complex samples, unsatisfactory edge detection and difficulties in identifying diverse samples. MFDS-Net uses DO-Conv to obtain better training results and lower floating point computation. MDPM brings a large amount of weak feature information into focus and makes an important contribution to the performance of the network. GSEM completes the correlation of feature information and enhances the strength of macro feature information, while DFIM completes the fusion of shallow and deep features to highlight the difference features. Later, we optimized the training process of the network through the deep supervision mechanism. MFDS-Net has effectively annotated the building changes in Texas between 5 and 14 years and its annotation accuracy is higher than the mainstream networks in the field of remote sensing change detection. Compared to mainstream networks, MFDS-Net has shown a higher ability to withstand environmental factors such as adversarial lighting and shadows in the annotation of buildings in Christchurch, New Zealand. Regarding the building changes in Guangzhou, China, MFDS-Net has demonstrated good performance in the detection of multi-colored changes. Our future work will focus on exploring more optimized and streamlined architectures based on MFDS-NET to address the ever-growing demands and challenges.

\bibliographystyle{IEEEtran}
\bibliography{ref}


\end{document}